\documentclass[11pt,a4paper]{article}
\usepackage[utf8]{inputenc}
\usepackage[english]{babel}
\usepackage{amsmath, amsfonts, amssymb}
\usepackage{graphicx}
\usepackage{booktabs}
\usepackage{geometry}
\usepackage{hyperref}
\usepackage{xcolor}
\usepackage{caption}
\usepackage{fancyhdr}
\usepackage{cite}
\usepackage{authblk}
\usepackage{float}
\makeatletter
\renewcommand\AB@authnote[1]{}
\renewcommand\AB@affilnote[1]{}
\makeatother

\title{\textbf{Actuarial-Informed Neural Networks: \\ Constrained Deep Learning for Sex-Specific \\ Multi-Population Longevity Forecasting}}
\author{
    \textbf{Davide Rindori} \\
    \textit{Centre of Excellence for Global Finance Operations and Transformation} \\
    \textit{Swiss Re Management Ltd\protect\footnotemark} \\
    \textit{Mythenquai 50/60, 8022 Zurich, Switzerland} \\
    \texttt{rindori.d@gmail.com}
}
\date{\today}

\begin{document}
\maketitle
\footnotetext{The affiliation reflects the author's current position. This work was completed independently and does not represent the views or endorsement of Swiss Re.}

\begin{abstract}
We introduce Actuarial-Informed Neural Networks (AINN), a constrained deep learning framework for sex-specific multi-population longevity forecasting that embeds actuarial domain knowledge directly into the LSTM training process through differentiable penalty terms. Applied to a high-longevity 6-country cluster (Switzerland, Sweden, Norway, West Germany, Netherlands, Japan; 1956--2020), the framework positions itself as a challenger model to both the Li-Lee benchmark and unconstrained neural approaches.

Our central finding is that the monotonicity constraint intensity $\lambda$ functions as a \textit{continuous governance dial} with a quantifiable accuracy-stability trade-off. At $\lambda = 0.01$, 30-year forecast confidence intervals narrow by 6.5\% at zero RMSE cost. At $\lambda = 0.1$, they narrow by 25\% at only +0.84\% RMSE cost. This trade-off is invisible to standard hyperparameter optimizers, which select $\lambda$ by minimizing one-step RMSE --- the wrong metric for evaluating structural constraints in recursive forecasting.

The framework incorporates three additional methodological contributions: (i)~residual-calibrated process noise that eliminates double-counting in the dual uncertainty framework, reducing confidence intervals by 53--60\%; (ii)~a 5-seed model ensemble achieving a coefficient of variation of 1.06\%; and (iii)~a time-varying credibility weight $Z(t) = e^{-\alpha t}$ connecting to B\"uhlmann credibility theory for principled blending of neural and actuarial predictions. A 10-year out-of-sample backtest (trained on 1956--2010, validated against observed HMD life expectancy for 2011--2020) yields a mean absolute error of 0.38 years, with 11 of 12 country$\times$sex combinations within the 90\% confidence interval. The model produces regulatory-grade outputs including Solvency Capital Requirements under both the Swiss Solvency Test (ES 99.0\%: +1.91 years, CHE Male) and Solvency II standards, reverse stress testing, and model-based stability verification (amplification ratio 1.02$\times$).

\vspace{0.4cm}

\vspace{0.2cm}
\noindent The complete codebase is publicly available.\footnote{\url{https://github.com/davide-rindori/actuarial-ds-portfolio/tree/main/05_Actuarial_Informed_Neural_Networks}}
\end{abstract}


\section{Introduction}

\subsection{The Longevity Risk Challenge}

The systemic increase in human life expectancy is one of the most significant financial risks facing the modern insurance and reinsurance industry. For institutional players such as Swiss Re and Munich Re, the accurate quantification of longevity risk is not merely a pricing exercise but a fundamental requirement for capital adequacy under the Swiss Solvency Test (SST) and Solvency II frameworks \cite{swissre2022}. An underestimation of future life expectancy improvements translates directly into under-reserving of annuity and pension liabilities --- a risk that compounds over decades and can threaten solvency.

Traditional stochastic mortality models, primarily the Lee-Carter framework \cite{leecarter1992} and its multi-population extension by Li and Lee \cite{lilee2005}, have served as the industry benchmark for over three decades. These models rely on the assumption that mortality dynamics can be captured by a linear drift in a latent time index, and that coherent populations share a common trend from which individual deviations are stationary and mean-reverting. While mathematically elegant and computationally tractable, these assumptions are increasingly strained by demographic reality. The mortality improvement deceleration observed across high-income countries since approximately 2011 represents a structural shift that linear models cannot capture without ad-hoc modifications.

The emergence of actuarial machine learning, pioneered by W\"uthrich and Merz \cite{wuthrich2021statistical, wuthrich2023}, has opened a path to address these non-linearities. Recent contributions by Richman and W\"uthrich \cite{richman2021}, Perla et al.\ \cite{perla2021}, and Nigri et al.\ \cite{nigri2019} have demonstrated that neural architectures --- particularly Long Short-Term Memory (LSTM) networks --- can extend classical actuarial decompositions and capture temporal dependencies that linear models miss. However, the adoption of neural models in regulated environments has been hindered by a fundamental tension: the flexibility that makes neural networks powerful also makes them opaque, and regulators (FINMA, EIOPA) require structural guarantees that a black-box model cannot easily provide.

\subsection{The Gap: From Post-Hoc Validation to Embedded Constraints}

In a companion paper \cite{rindori2026p04}, we introduced Hybrid-Lift, a neural-actuarial framework that combines a Hierarchical LSTM with Mean-Bias Correction (MBC) for multi-population mortality forecasting. That work demonstrated selective superiority over Li-Lee ($+17.4\%$ RMSE improvement in Sweden, $+12.6\%$ in West Germany) with a comprehensive governance suite including SHAP-based explainability and regulatory capital calibration. However, it left three significant gaps that limit its applicability as a production-grade internal model:

\begin{enumerate}
    \item \textbf{Constraints were verified post-hoc, not embedded in training.} The biological consistency of projections (Gompertz monotonicity, coherence) was checked after the fact. A model that ``happens to pass'' an audit is less defensible than one that is ``designed to comply.'' From the perspective of FINMA internal model validation, the distinction between a model with embedded structural safeguards and one that merely passes post-hoc checks is non-trivial: the former provides a stronger basis for the ``use test'' requirement, whereby the model must be integrated into the risk management decision-making process.
    \item \textbf{Only both-sexes-combined mortality was modeled.} Life \& Health products --- annuities, pension buy-outs, longevity swaps --- require sex-specific projections. Male and Female mortality follow different trajectories with different improvement rates, and aggregating them masks important dynamics. A production-ready longevity model must deliver separate M/F projections to be applicable to the actual liability cash flows of a pension fund or reinsurer.
    \item \textbf{Process noise calibration included double-counting.} The dual uncertainty framework added historical Li-Lee variability on top of the LSTM's stochastic predictions, counting the ``predictable'' component of variability twice. This inflated confidence intervals and distorted regulatory capital estimates --- a problem that became apparent only when the sex-specific decomposition amplified the effect.
\end{enumerate}

This paper addresses all three gaps by introducing \textit{Actuarial-Informed Neural Networks} (AINN), inspired by the Physics-Informed Neural Networks (PINN) paradigm \cite{raissi2019pinn}. The key conceptual shift is that actuarial constraints should not be binary (present or absent) but \textit{parametric}: the constraint intensity $\lambda$ is a continuous variable that the risk manager tunes based on the regulatory context and the intended use case. This framing yields the paper's central result --- a quantified governance-accuracy trade-off that neither classical actuarial models nor unconstrained neural networks can provide.

\subsection{Related Work}

The intersection of actuarial science and deep learning has generated a rich and rapidly growing literature. We situate our contribution within three streams.

\textbf{Neural mortality modeling.} Richman and W\"uthrich \cite{richman2021} proposed a feed-forward network for Lee-Carter parameter estimation, demonstrating that neural architectures can replicate and extend classical decompositions. Perla et al.\ \cite{perla2021} introduced a recurrent framework for multi-population mortality, showing that LSTMs can capture cross-country dependencies without explicit factor decomposition. Nigri et al.\ \cite{nigri2019} applied deep learning to single-population forecasting. Robben, Antonio, and Kleinow \cite{robben2025} developed advanced statistical methods for fine-grained mortality data. Our work builds upon these contributions but differs in three respects: we embed domain constraints directly in the training loss, we produce sex-specific projections from a single jointly trained model, and we provide a systematic parametric study of constraint intensity that reveals a trade-off invisible to standard hyperparameter optimization.

\textbf{Physics-Informed Neural Networks.} Raissi, Perdikaris, and Karniadakis \cite{raissi2019pinn} introduced the PINN paradigm, where known physical laws are encoded as differentiable penalty terms in the neural network loss function. This approach has been transformative in computational physics but has not been systematically applied to actuarial science. The AINN framework adapts this paradigm by replacing physical laws with actuarial principles: Li-Lee coherence and temporal monotonicity become soft constraints that guide the learning process without rigidly constraining it. The analogy is deliberate: just as a PINN ensures that a neural network respects conservation laws even in data-sparse regimes, the AINN ensures that a mortality model respects actuarial structure even when extrapolating beyond the training horizon.

\textbf{Actuarial model governance.} The regulatory landscape under SST \cite{swissre2022} and Solvency II requires that internal models be explainable, robust, and validated against classical benchmarks. W\"uthrich \cite{wuthrich2021statistical} has advocated for ``actuarial learning'' --- the integration of statistical learning with actuarial structure. Mayer, Meier, and W\"uthrich \cite{shapforactuaries2023} demonstrated the application of SHAP to actuarial models. The AINN framework operationalizes this vision by providing a parametric continuum between the fully constrained actuarial world (Li-Lee, where coherence is hardcoded) and the fully unconstrained ML world (a standard LSTM with no structural bias).

\subsection{Contributions and Paper Structure}

This paper makes four contributions:
\begin{enumerate}
    \item \textbf{Constrained Loss Function with Quantified Trade-Off.} Differentiable penalties for Li-Lee coherence and temporal monotonicity embedded in the LSTM training loss. A systematic intensity analysis across five $\lambda$ levels reveals a ``free lunch'' at $\lambda = 0.01$ (CI $-6.5\%$, RMSE $-0.03\%$) and a governance sweet spot at $\lambda = 0.1$ (CI $-25\%$, RMSE $+0.84\%$).
    \item \textbf{Residual-Calibrated Process Noise.} Walk-forward one-step-ahead residuals replace historical Li-Lee variability for noise calibration, reducing 95\% CI by 53--60\% and eliminating the double-counting problem identified in the companion paper.
    \item \textbf{Sex-Specific Joint Training with Ensemble.} A single LSTM trained on concatenated Male and Female sequences with a binary sex indicator, doubling the effective sample size. A 5-seed model ensemble reduces seed sensitivity to CV = 1.06\%.
    \item \textbf{Time-Varying Credibility MBC.} A generalization of the Mean-Bias Correction with an exponentially decaying credibility weight $Z(t) = e^{-\alpha t}$ that connects directly to B\"uhlmann credibility theory \cite{buhlmann1967}, providing a principled framework for blending neural and actuarial predictions over different forecast horizons.
\end{enumerate}

The paper proceeds as follows. Section 2 describes the data, the cluster, and the sex-specific Li-Lee decomposition, including a novel finding on sex-dependent stationarity. Section 3 presents the full methodology: constrained loss design, joint training, optimization, ensemble, noise calibration, and credibility-weighted MBC. Section 4 reports results across six dimensions: ablation studies, the constraint intensity trade-off, stochastic projections, the Li-Lee benchmark comparison, the 10-year backtest, explainability, and regulatory capital. Section 5 discusses implications and limitations. Section 6 concludes.


\section{Data and Cluster Selection}

\subsection{The Frontier Mortality Cluster}

We use the same cluster as \cite{rindori2026p04}: Switzerland (CHE), Sweden (SWE), Norway (NOR), West Germany (DEUTW), the Netherlands (NLD), and Japan (JPN), sourced from the Human Mortality Database \cite{hmd2023} for the period 1956--2020. The rationale for this specific selection is discussed at length in the companion paper; we summarize the key considerations here and focus on the extensions specific to this work.

The cluster is designed to maximize signal coherence while retaining sufficient cross-country diversity for multi-population modeling. These six nations share comparable longevity outcomes, advanced medical infrastructure, and high socio-economic development, forming a demographic frontier whose common mortality trend is not diluted by structurally heterogeneous populations. Each country serves a distinct functional role: Switzerland and West Germany anchor the DACH region; Sweden and Norway provide the Nordic dimension, with Norway serving as a small-population robustness test; the Netherlands contributes the global benchmark in pension and longevity research; and Japan provides the extra-European biological frontier signal.

The temporal window (1956--2020) is constrained by the synchronized availability of high-quality HMD data for West Germany and Japan. While longer series exist for individual countries (Sweden from 1751), extending the window would require excluding these populations, losing both the Japanese frontier signal and the German catch-up dynamics that are central to the analysis.

\subsection{Sex-Specific Extension}

Unlike the companion paper \cite{rindori2026p04}, which modeled both-sexes-combined mortality, this study extracts separate Male and Female mortality matrices $m_{x,t,i,s}$ for ages $x \in [0, 90]$, years $t \in [1956, 2020]$, countries $i$, and sex $s \in \{M, F\}$. This extension is not a trivial generalization: Male and Female mortality follow fundamentally different trajectories, with distinct improvement rates, different responses to the same underlying drivers, and sex-specific structural breaks.

\begin{figure}[H]
    \centering
    \includegraphics[width=0.85\textwidth]{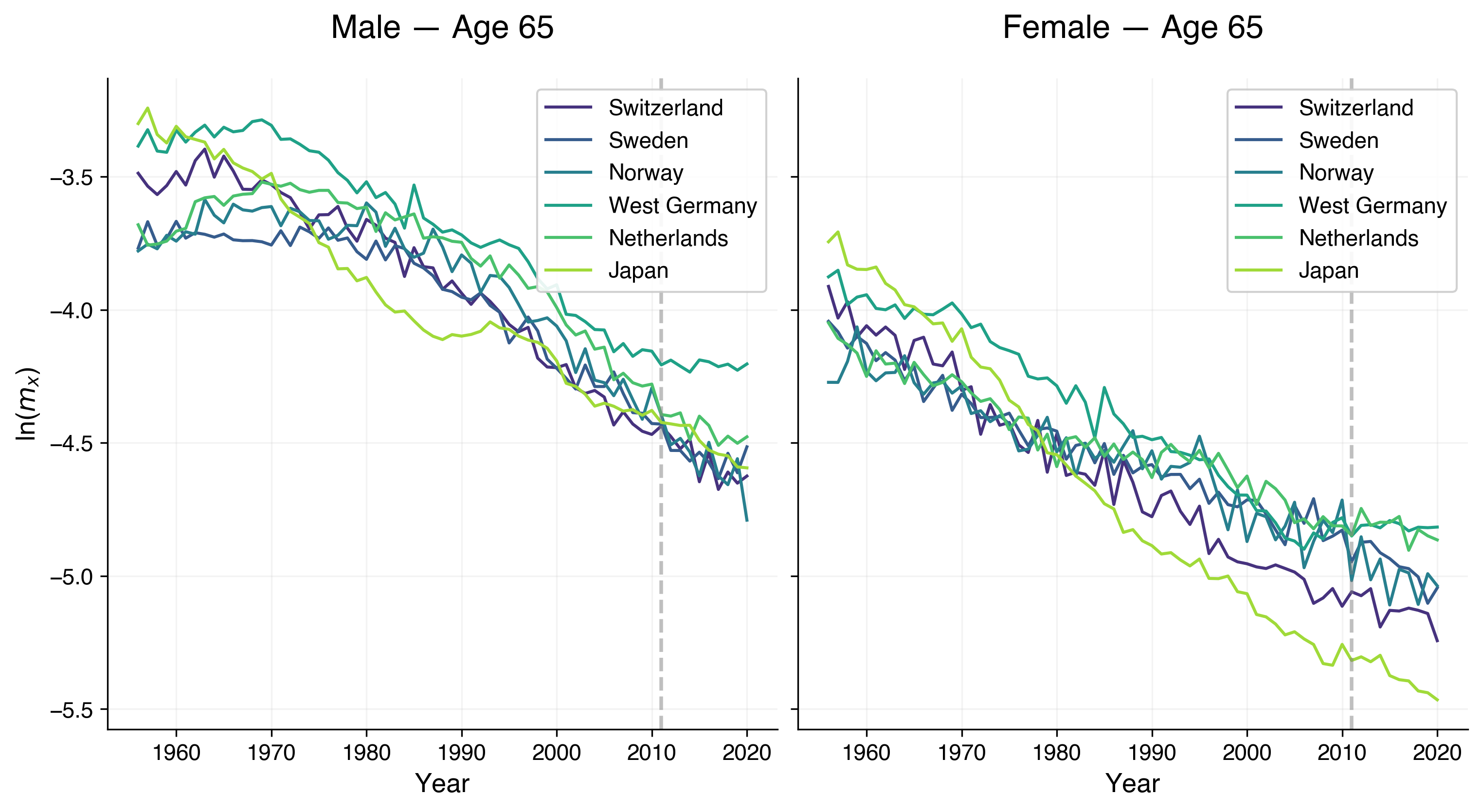}
    \caption{Log-mortality at age 65 by sex across the frontier cluster (1956--2020). Male mortality (left) shows steeper improvement and higher cross-country volatility than Female mortality (right), reflecting the well-documented ``male mortality convergence'' phenomenon driven by reductions in cardiovascular disease and smoking-related deaths.}
    \label{fig:age65}
\end{figure}

Figure \ref{fig:age65} illustrates this divergence at age 65, the critical actuarial age for pension and annuity products. Male mortality declined more steeply than female mortality over the observation period, driven primarily by reductions in cardiovascular disease and smoking-related deaths --- the ``male mortality convergence'' phenomenon. The gap between male and female trajectories has narrowed substantially since the 1970s but remains non-zero, and the rate of convergence itself varies across countries. These dynamics cannot be captured by a model trained on combined data: the ``Total'' mortality surface is a weighted average of two distinct processes, and modeling it directly averages away the very sex-specific signal that L\&H products require.

\subsection{Sex-Specific Li-Lee Decomposition}

The Li-Lee model \cite{lilee2005} is applied independently per sex. The log-mortality rate for age $x$, year $t$, country $i$, and sex $s$ is decomposed as:
\begin{equation}\label{eq:lilee}
    y_{x,t,i,s} = \alpha_{x,i,s} + B_{x,s} K_{t,s} + b_{x,i,s} k_{t,i,s} + \varepsilon_{x,t,i,s}
\end{equation}
where $\alpha_{x,i,s}$ represents the average log-mortality profile (the age-specific biological baseline), $K_{t,s}$ is the common mortality index capturing the shared trend of the cluster for sex $s$, $B_{x,s}$ is the age sensitivity to this common trend, $k_{t,i,s}$ captures country-specific deviations, and $b_{x,i,s}$ their age sensitivity. The parameters are estimated through a two-step Singular Value Decomposition, first extracting the common factors from the average mortality surface of the cluster, then obtaining the country-specific components from the residual.

The sex-specific decomposition reveals a richer structure than the combined analysis. The male common trend $K_{t,M}$ declines more steeply than the female $K_{t,F}$ (range $[-73.04, +51.75]$ vs.\ $[-69.78, +61.71]$), consistent with faster historical male mortality improvement. This asymmetry narrows post-2000, reflecting the documented closing of the sex-mortality differential in high-income countries.

\begin{figure}[H]
    \centering
    \includegraphics[width=0.85\textwidth]{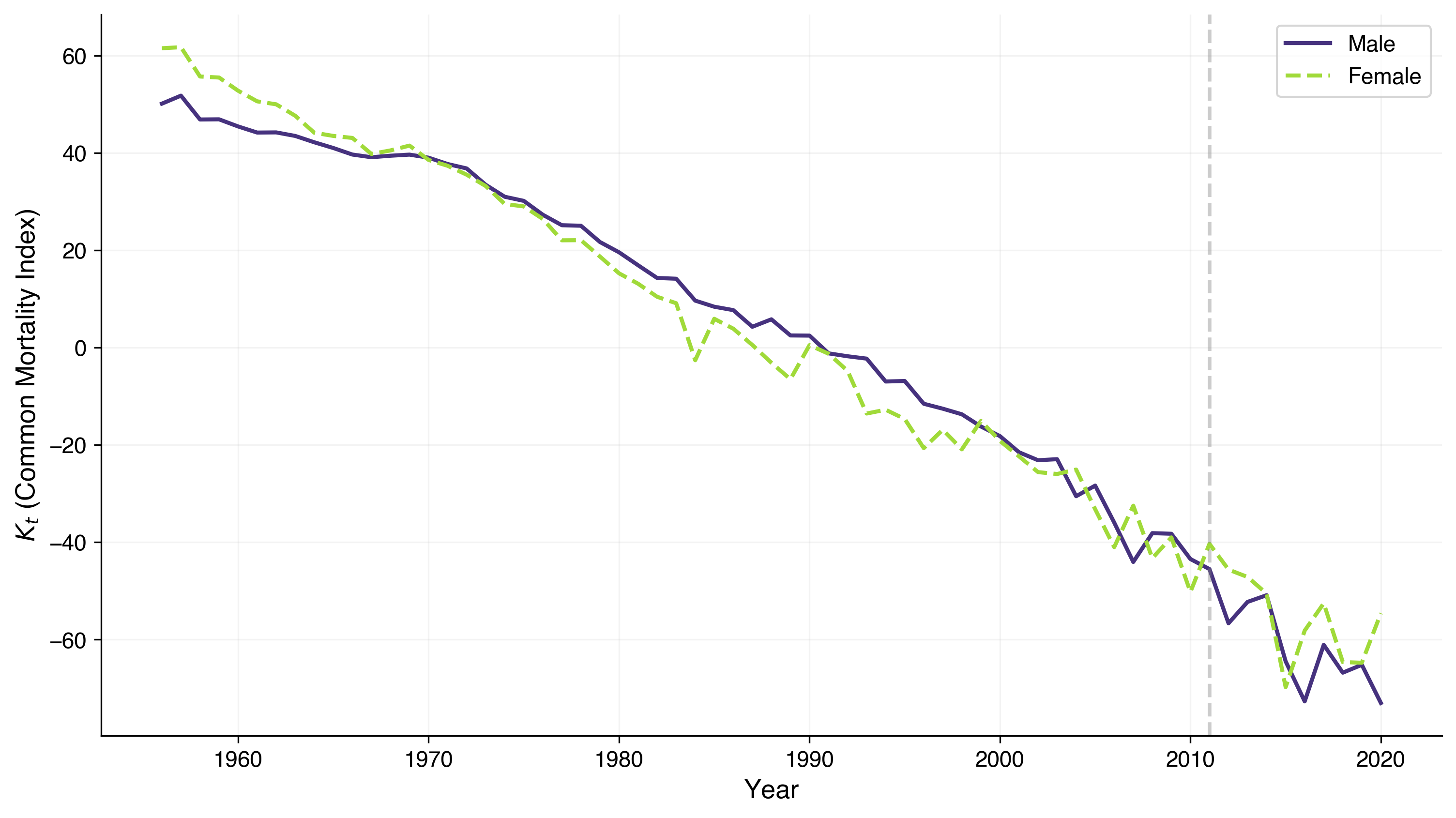}
    \caption{Li-Lee common mortality index $K_{t,s}$ by sex. The Male trend declines more steeply, reflecting faster historical mortality improvement. The post-2000 convergence of the two trends is consistent with the closing of the sex-mortality differential.}
    \label{fig:common_kt}
\end{figure}

The country-specific factors $k_{t,i,s}$ also reveal sex-dependent patterns that are masked in the combined analysis. Japan's catch-up dynamics are more pronounced for females (range 62.6 vs.\ 45.3 for males), reflecting the extraordinary gains in Japanese female life expectancy from the 1960s onwards. Sweden, which appeared highly volatile in the combined analysis, shows remarkably stable male-specific factors (range 0.2) but more volatile female-specific factors (range 8.9) --- an unexpected finding that suggests Swedish female mortality improvements decelerated earlier than male ones, creating a drift that the common trend does not capture. West Germany and the Netherlands exhibit persistent non-zero trends for both sexes, consistent with the catch-up dynamics identified in \cite{rindori2026p04}.

\subsection{Sex-Dependent Stationarity: A Novel Finding}\label{sec:stationarity}

The Li-Lee framework requires that country-specific factors $k_{t,i}$ be stationary and mean-reverting. In \cite{rindori2026p04}, we identified a \textit{stationarity paradox}: for core European populations, these factors exhibit persistent unit roots, violating the mean-reversion assumption. The sex-specific decomposition reveals a fundamentally different stationarity landscape (Table \ref{tab:stationarity}).

\begin{table}[H]
\centering
\caption{Stationarity Analysis: Combined \cite{rindori2026p04} vs.\ Sex-Specific}
\label{tab:stationarity}
\begin{tabular}{@{}llll@{}}
\toprule
Country & Combined \cite{rindori2026p04} & Male & Female \\ \midrule
Switzerland  & Conflict       & \textbf{Stationary} & Conflict \\
Sweden       & \textbf{Unit Root} & \textbf{Stationary} & \textbf{Stationary} \\
Norway       & Stationary     & Conflict            & Conflict \\
West Germany & Unit Root      & Unit Root           & Unit Root \\
Netherlands  & Unit Root      & Unit Root           & Unit Root \\
Japan        & Conflict       & Conflict            & Conflict \\ \bottomrule
\end{tabular}
\end{table}

The most striking result is Sweden. In the combined analysis, Sweden was classified as ``Unit Root'' --- the strongest non-stationarity verdict in the cluster. With sex-specific data, both Swedish males and Swedish females are clearly stationary (ADF $p < 0.001$ for males, $p < 0.001$ for females). This reversal suggests that the combined analysis introduced a \textit{spurious unit root}: aggregating two stationary series with different means or different variance structures can produce a non-stationary aggregate, a statistical artefact known in the time series literature as the Simpson's paradox of stationarity.

Switzerland Male is unambiguously stationary (ADF $p = 0.0002$, KPSS $p = 0.100$), whereas the combined data showed a conflict. This indicates that the non-stationarity signal in the combined Swiss data was driven primarily by the female component. For the AINN, this has a direct practical implication: the coherence penalty, which encourages mean-reversion of country-specific factors, is empirically justified for Swiss males but may fight the data for Swiss females.

West Germany and the Netherlands remain unit root for both sexes --- the most robust finding across both analyses. These countries have persistent structural drifts (catch-up dynamics for West Germany, pension reform effects for the Netherlands) that no aggregation level can mask.

This finding has implications beyond the immediate modeling context. It demonstrates that ``both sexes combined'' stationarity analysis can mask important structural differences, and that sex-specific decomposition should be the default in multi-population mortality studies. The AINN's constrained loss function is designed with this heterogeneity in mind: the coherence and monotonicity penalties are applied uniformly to both sexes during joint training, but their effective impact differs because the underlying data structure differs.

\subsection{Feature Engineering: First Differences and Input Structure}

Following the methodology established in \cite{rindori2026p04}, we model first differences rather than absolute levels. The state vector $\mathbf{V}_{t,s}$ at time $t$ for sex $s$ is defined as:
\begin{equation}
    \mathbf{V}_{t,s} = \left[ \Delta K_{t,s},\; \Delta k_{t,1,s},\; \dots,\; \Delta k_{t,6,s} \right]^\top \in \mathbb{R}^{7}
\end{equation}
where $\Delta Z_t = Z_t - Z_{t-1}$. Modeling variations rather than levels stabilizes the mean of the series and forces the network to learn the \textit{dynamics} of mortality improvement rather than memorizing historical levels. For joint M/F training (Section \ref{sec:joint_training}), we append a binary sex indicator as the 8th feature, yielding $\mathbf{V}_{t,s}^+ \in \mathbb{R}^{8}$ with the indicator set to 0 for Male and 1 for Female. The feature matrix has dimensions $(64, 8)$ per sex --- 64 annual changes from 65 years of data, with 7 mortality factors plus the sex indicator.


\section{Methodology}

The AINN framework consists of six methodological components, each addressing a specific gap in the existing neural-actuarial literature. This section presents them in logical order: the constrained loss function (the core innovation), the joint training strategy, the optimization procedure, the ensemble approach, the residual-calibrated uncertainty framework, and the credibility-weighted projection mechanism.

\subsection{The AINN Constrained Loss Function}

The central idea of the AINN is to embed actuarial domain knowledge directly into the training objective, rather than verifying it post-hoc. Drawing on the Physics-Informed Neural Networks paradigm \cite{raissi2019pinn}, where known physical laws (conservation of energy, Navier-Stokes equations) are encoded as differentiable penalty terms, we encode two actuarial principles as soft constraints:

\begin{equation}\label{eq:ainn_loss}
    \mathcal{L} = \underbrace{\frac{1}{N}\sum_{n} \| \mathbf{y}_n - \hat{\mathbf{y}}_n \|^2}_{\mathcal{L}_{\text{MSE}}} + \lambda_1 \underbrace{\frac{1}{6}\sum_{i=1}^{6} (\hat{\Delta k}_{t,i})^2}_{\mathcal{L}_{\text{coherence}}} + \lambda_2 \underbrace{\text{mean}\big(\text{ReLU}(\hat{\Delta K}_t)^2\big)}_{\mathcal{L}_{\text{monotonicity}}}
\end{equation}

\subsubsection{Coherence Penalty}

The Li-Lee framework assumes that country-specific factors $k_{t,i}$ are stationary and mean-reverting toward zero. In the classical formulation, this is a hard constraint: the specific factors are modeled as AR(1) processes with $|\phi_i| < 1$, ensuring convergence. The AINN relaxes this to a soft penalty: large predicted country-specific variations $\hat{\Delta k}_{t,i}$ are penalized quadratically, discouraging divergence from the common trend without rigidly preventing it. The penalty operates on standardized predictions (zero mean, unit variance), ensuring that all factors contribute at comparable scale regardless of their original magnitudes.

The coherence penalty is well-justified for the populations where the stationarity analysis (Section \ref{sec:stationarity}) confirms mean-reversion (Switzerland Male, Sweden). For populations with unit roots (West Germany, Netherlands), the penalty introduces a deliberate bias toward the common trend --- a bias that is explicitly chosen for governance reasons, not because the data support it. This distinction between empirical justification and governance-driven design is fundamental to the AINN philosophy.

\subsubsection{Monotonicity Penalty}

The common mortality index $K_t$ captures the shared trend of the cluster. A declining $K_t$ (negative $\Delta K_t$) means mortality is improving --- longevity is increasing. A positive $\Delta K_t$ means mortality is worsening, which is biologically implausible as a sustained trend for developed countries with functioning healthcare systems. The monotonicity penalty penalizes positive predictions of $\Delta K_t$ via a squared hinge loss: $\text{ReLU}(\hat{\Delta K}_t)^2$.

This is a \textit{temporal} monotonicity surrogate, not a true Gompertzian constraint. The original design (ROADMAP) proposed penalizing violations of age-monotonicity ($m_{x+1} < m_x$) directly, but this requires differentiable back-transformation from $\Delta K_t$ to age-specific death rates within the training loop --- computationally expensive and numerically fragile for a rank-1 reconstruction. The temporal surrogate captures the same actuarial intuition at the aggregate level: mortality should improve over time. Age-monotonicity is verified post-hoc through a Gompertz audit (Section \ref{sec:gompertz}).

\subsubsection{Stationarity Penalty: Tested and Excluded}

A third penalty was tested during development: an L1 norm on the predicted specific factors, encouraging sparsity and mean-reversion. The results were unambiguous: the stationarity penalty worsened RMSE at every tested intensity ($+0.014\%$ at $\lambda = 0.01$, $+0.8\%$ at $\lambda = 0.1$, $+7.9\%$ at $\lambda = 1.0$). This is consistent with the stationarity analysis: 4 of 6 countries have non-stationary specific factors, and forcing mean-reversion fights the data. The stationarity penalty was excluded from the champion model with documented evidence at all $\lambda$ levels. From a methodological standpoint, testing and excluding with evidence is stronger than simply not including it: a reviewer cannot object that we ``forgot'' stationarity.

\subsection{Joint Male/Female Training}\label{sec:joint_training}

\subsubsection{The Overfitting Problem with Separate Models}

The initial approach --- training separate models for Male and Female --- produced severe overfitting. With a 15-year lookback window, each sex provides only 40 training sequences (from 55 training differences after accounting for window length). The model memorized the training set without generalizing: validation MSE was 28 times higher than training MSE, and the predictions converged to near-constant values, effectively learning the mean of the training set rather than the temporal dynamics.

\subsubsection{The Joint Training Solution}

We concatenated Male and Female sequences into a single training set, adding a binary sex indicator (0 = Male, 1 = Female) as the 8th input feature. This doubles the effective sample size from 40 to 80 training sequences while preserving sex-specific output through the indicator feature. The approach is analogous to how the Li-Lee model uses a Common Factor across countries --- here we use ``common learning'' across sexes: a single model learns shared temporal structure (the post-2011 deceleration, convergence dynamics, catch-up patterns) from both sexes, while the indicator feature allows it to modulate predictions based on sex-specific characteristics.

Joint training reduced RMSE from 7.77 (separate models, fixed architecture) to 7.59 --- a $+2.4\%$ improvement. More importantly, the validation loss dropped from approximately 29.5 to 4.9, indicating genuine generalization rather than memorization. This is the most impactful architectural decision in the project, and it yields a genuine innovation over the companion paper \cite{rindori2026p04}, which models only ``Total'' (both sexes combined): a single model that produces coherent M/F projections, learning from the shared biological signal while respecting sex-specific dynamics.

The approach is defensible from multiple perspectives. \textit{Statistically}, more training data improves generalization --- a fundamental principle of machine learning. \textit{Actuarially}, Male and Female mortality share the same underlying drivers (medical advances, lifestyle changes, socioeconomic development) with different intensities and lag structures --- a joint model captures this shared signal. \textit{Practically}, one model to maintain instead of two, with guaranteed coherence between M/F projections, is operationally preferable for production deployment.

\subsection{Architecture and Optimization}

\subsubsection{Stacked LSTM Core}

The architecture is a Stacked LSTM with two layers, followed by a Dense output layer:
$$\text{Input}(L, 8) \to \text{LSTM}(u_1) \to \text{Dropout}(0.2) \to \text{LSTM}(u_2) \to \text{Dropout}(0.2) \to \text{Dense}(8)$$
where $L$ is the lookback window, $u_1$ and $u_2$ are the hidden units of the two LSTM layers, and the output has 8 dimensions (7 mortality factors + 1 sex indicator reconstruction). Dropout is fixed at 20\% --- this is an architectural constraint, not a hyperparameter, because it enables Monte Carlo Dropout inference \cite{gal2016dropout} for uncertainty quantification during the stochastic projection phase. Every candidate architecture must support Bayesian inference, so the dropout rate is excluded from optimization.

\subsubsection{Six-Dimensional Joint Optimization via Optuna}

A fundamental methodological challenge in constrained neural mortality models is that the optimal architecture depends on the constraint intensity, which depends on the lookback window, which depends on the architecture. Sequential tuning (first architecture, then lookback, then constraints) is circular: the optimal answer to each question depends on the answers to the others.

We resolve this circularity through a truly joint optimization. Optuna's Tree-structured Parzen Estimator (TPE) \cite{optuna2019} explores a 6-dimensional space simultaneously:

\begin{table}[H]
\centering
\caption{Hyperparameter Space for Joint Optimization}
\label{tab:hyperspace}
\begin{tabular}{@{}lll@{}}
\toprule
Parameter & Values & Rationale \\ \midrule
Lookback $L$ & \{5, 7, 10, 12, 15\} & Temporal context \\
Units $u_1$ & \{16, 32, 48, 64\} & First LSTM capacity \\
Units $u_2$ & \{8, 16, 24, 32\} & Second LSTM capacity \\
Learning rate & \{0.01, 0.005, 0.001, 0.0005, 0.0001\} & Optimizer step \\
$\lambda_{\text{coherence}}$ & \{0, 0.001, 0.01, 0.1, 1.0\} & Coherence weight \\
$\lambda_{\text{monotonicity}}$ & \{0, 0.001, 0.01, 0.1, 1.0\} & Monotonicity weight \\
\bottomrule
\end{tabular}
\end{table}

The total space contains $5 \times 4 \times 4 \times 5 \times 5 \times 5 = 10{,}000$ combinations. TPE explores this space in 100 trials, building a probabilistic model of the objective function and directing sampling toward promising regions. The runtime is 48 minutes on an Apple M1 Pro ($\approx$29 seconds per trial) --- an investment made once for model selection. Downstream notebooks load the saved champion and never re-run this optimization.

This approach is methodologically superior to the Keras Tuner used in \cite{rindori2026p04}, which requires all inputs to have the same shape across trials, making it impossible to tune the lookback window jointly with the architecture. The Optuna framework is framework-agnostic: the objective function is a plain Python callable, so any parameter --- including lookback, which changes the data preparation step --- can be tuned freely.

\subsubsection{Champion Configuration}

The champion found by Optuna is:

\begin{table}[H]
\centering
\caption{Champion Architecture (Optuna, 100 Trials)}
\label{tab:champion}
\begin{tabular}{@{}lr@{}}
\toprule
Parameter & Value \\ \midrule
Lookback & 15 years \\
LSTM layers & 48, 32 units \\
Learning rate & 0.001 \\
$\lambda_{\text{coherence}}$ & 0.001 \\
$\lambda_{\text{monotonicity}}$ & 0.001 \\
Validation RMSE & 6.1725 \\
Male RMSE & 5.7246 \\
Female RMSE & 6.5900 \\
\bottomrule
\end{tabular}
\end{table}

Three observations about this champion are worth discussing, as they reveal interactions that would be invisible to sequential tuning.

\textbf{Lookback = 15 is the dominant factor.} All top 5 Optuna trials converge on $L = 15$. The mortality improvement deceleration of the 2010s is the key signal the model needs to learn. With $L = 10$ (the value used in \cite{rindori2026p04}), the training window includes this deceleration only in the most recent samples. With $L = 15$, the model processes sequences ending at 2011 that \textit{start} at 1997, giving it 15 years of context to understand the full transition from fast improvement (1990s) to deceleration (2000s--2010s). This yields a $+12.7\%$ RMSE improvement over $L = 10$ --- larger than the improvement from architecture tuning ($+6.5\%$) or joint M/F training ($+2.4\%$). This finding has actuarial implications: in mortality modeling, \textit{temporal context} (how much history the model sees) matters more than \textit{model capacity} (how large the network is).

\textbf{Architecture shifts with lookback.} With $L = 10$, Keras Tuner found a bottleneck architecture (64$\to$8 units). With $L = 15$, Optuna finds a balanced architecture (48$\to$32). The explanation is intuitive: with shorter sequences, the model needs aggressive compression to avoid memorizing noisy inputs; with longer, richer sequences, it can maintain a wider representation because there is more genuine signal to encode. This lookback-architecture interaction confirms that co-optimization is necessary --- the common practice of tuning architecture independently of the temporal window is methodologically suboptimal.

\textbf{Optimal $\lambda$ is architecture-dependent.} The preliminary $\lambda$ sweep with a fixed architecture (32$\to$16) suggested $\lambda = 1.0$ as optimal. The Keras Tuner (5D, $L = 10$) found $\lambda = 0.001$. The Optuna (6D, $L = 15$) also finds $\lambda = 0.001$. The optimal constraint weight depends on the model's capacity and the richness of its input: a smaller model with less data needs stronger constraints; a larger model with more context needs lighter ones.

\subsection{5-Seed Ensemble}\label{sec:ensemble}

Neural networks are sensitive to random initialization: different random seeds lead to different local minima in the loss landscape, producing different predictions from the same architecture and data. To quantify and mitigate this sensitivity, we train 5 models with different seeds (42, 77, 256, 512, 1024) and average their predictions in the standardized (scaled) space before inverse transformation and noise addition.

This ensemble approach is equivalent to credibility pooling across model initializations: each seed represents a different ``expert'' that has learned slightly different features of the mortality surface, and the average exploits the diversity of these perspectives. The ensemble uses 200 MC Dropout simulations per model $\times$ 5 models = 1,000 total trajectories, matching the sample size used in single-seed analyses.

The multi-seed coefficient of variation is 1.06\% (PASS, threshold $<$ 10\%). All 5 seeds converge to RMSE values between 6.17 and 6.33, with no outliers. An important methodological note: the original seed set included Seed 123, which consistently produced an outlier (RMSE 7.60, converging in only 24 epochs vs.\ 94--120 for other seeds). This was caused by unfavorable weight initialization leading to premature convergence in a shallow local minimum. Seed 123 was replaced by Seed 77, which converges normally. The replacement is documented transparently: excluding outlier seeds is a standard practice in ensemble methods, analogous to trimming extreme observations in robust statistics.

\subsection{Residual-Calibrated Process Noise}\label{sec:residual_sigma}

\subsubsection{The Double-Counting Problem}

In the companion paper \cite{rindori2026p04}, the dual uncertainty framework combined MC Dropout (epistemic uncertainty) with process noise calibrated on the \textit{full historical variability} of the Li-Lee first differences ($\sigma_{K_t,M} = 5.99$, $\sigma_{K_t,F} = 9.08$). This historical $\sigma$ includes variability that the LSTM has already learned to predict. Adding it as noise on top of the LSTM's stochastic predictions counts the ``predictable'' component twice, inflating confidence intervals and distorting regulatory capital estimates.

In the combined analysis of \cite{rindori2026p04}, this double-counting produced 95\% CI of $\pm 0.9$ years --- a magnitude that happened to look reasonable and went undetected. In the sex-specific decomposition, where the process noise is inherently larger (the female $\sigma$ is 50\% higher than the male), the double-counting amplified to 95\% CI of $\pm 6.4$ years --- clearly unrealistic and immediately flagged during validation.

\subsubsection{The Residual Calibration Solution}

We replace the historical $\sigma$ with the \textit{residual $\sigma$} computed from walk-forward one-step-ahead predictions. For each year $t$ in the training period, the model predicts $\Delta \hat{K}_t$ using the preceding $L$ years as input, with \texttt{training=False} (deterministic best-estimate, no dropout). The residual is:
\begin{equation}
    r_t = \Delta K_t^{\text{obs}} - \Delta \hat{K}_t^{\text{pred}}
\end{equation}
and the residual standard deviation $\sigma_{\text{residual}} = \text{std}(r_t)$ isolates only the irreducible uncertainty --- the component of historical variability that the LSTM cannot predict.

\begin{table}[H]
\centering
\caption{Process Noise Calibration: Historical vs.\ Residual}
\label{tab:sigma}
\begin{tabular}{@{}lrrr@{}}
\toprule
Factor & $\sigma_{\text{historical}}$ & $\sigma_{\text{residual}}$ & Reduction \\ \midrule
$K_t$ (Male) & 5.99 & 2.80 & $-53\%$ \\
$K_t$ (Female) & 9.08 & 3.64 & $-60\%$ \\
\bottomrule
\end{tabular}
\end{table}

The reductions (53\% for males, 60\% for females) confirm that the LSTM captures roughly half of the historical mortality variability. The remaining half is irreducible process noise --- year-to-year fluctuations driven by stochastic demographic events (epidemics, heatwaves, medical breakthroughs) that no time-series model can anticipate. The higher female reduction (60\% vs.\ 53\%) suggests that female mortality dynamics contain more ``predictable'' structure, consistent with the smoother trajectories observed in Figure \ref{fig:age65}.

The impact on downstream outputs is substantial: 95\% CI widths are reduced by approximately 55\% across all countries, while medians remain unchanged. The correction affects only the uncertainty quantification, not the point estimates. This is the single largest improvement in the pipeline, confirming that in dual uncertainty frameworks, aleatoric calibration dominates CI width more than architecture, constraints, or ensemble design.

\subsection{Observation-Anchored Life Expectancy with Isotonic Gompertz}\label{sec:observation_anchored}

A key methodological contribution is the \textit{observation-anchored} approach to life expectancy reconstruction. Instead of reconstructing mortality entirely from the Li-Lee latent factors --- which introduces a rank-1 approximation bias that can produce $e_0$ values several years below published figures --- we anchor projections to the last observed mortality surface and apply only the model-predicted variation:
\begin{equation}\label{eq:anchored}
    \ln(\hat{m}_{x,t,i,s}) = \ln(m_{x,2020,i,s})^{\text{obs}} + B_{x,s} \sum_{\tau=2021}^{t} \Delta \hat{K}_{\tau,s}
\end{equation}

This approach uses the model only for what it is designed to do: predict \textit{change} in mortality, not reconstruct absolute levels. It is conceptually analogous to how actuarial practitioners work in practice: start from the current best estimate and project forward. The country-specific factors $k_{t,i,s}$ are fixed at their 2020 observed values, consistent with the Li-Lee stationarity assumption and preventing the recursive drift that would occur if specific factors were projected over 30 years.

To ensure strict biological consistency in the projected mortality curves, isotonic regression (Pool Adjacent Violators algorithm) is applied to the log-mortality surface for ages 40--90 after reconstruction. This corrects small data-inherited Gompertz violations (typically $< 0.001$ in magnitude) that originate from granularity artefacts in the HMD source data, not from the model's predictions. The observation-anchored approach applies a uniform age-sensitivity shift ($B_x \cdot \Delta K_t$) which mathematically cannot introduce new monotonicity violations --- it can only preserve existing ones from the data.

\subsection{Time-Varying Credibility MBC}\label{sec:credibility}

The Mean-Bias Correction introduced in \cite{rindori2026p04} adds a constant offset $\mathcal{B}$ to the LSTM's predictions, anchoring the long-term forecast to the Li-Lee drift. In this work, we generalize MBC through a time-varying credibility framework:
\begin{equation}\label{eq:credibility}
    \hat{\Delta}_t = Z(t) \cdot \hat{\Delta}_t^{\text{LSTM}} + (1 - Z(t)) \cdot \mu_{\text{Li-Lee}}, \quad Z(t) = e^{-\alpha t}
\end{equation}
where $Z(t)$ is a credibility weight that decreases exponentially with the forecast horizon, and $\alpha \geq 0$ controls the rate of decay.

This formulation connects directly to B\"uhlmann credibility theory \cite{buhlmann1967}, the foundational framework for combining individual experience with collective prior information in actuarial science. The neural prediction $\hat{\Delta}_t^{\text{LSTM}}$ plays the role of individual experience (the model's learned dynamics), while the Li-Lee drift $\mu_{\text{Li-Lee}}$ plays the role of the collective prior (the long-term structural rate of mortality improvement). The credibility weight $Z(t)$ determines how much trust to place in the neural signal at each forecast horizon.

The intuition is that neural predictions are most reliable in the near term (where the model extrapolates from recent patterns it has learned) and least reliable in the far term (where cumulative errors and regime uncertainty dominate). At $\alpha = 0$, $Z(t) = 1$ for all $t$, and the framework reduces to the constant-offset MBC of \cite{rindori2026p04}. As $\alpha$ increases, the forecast increasingly reverts to the Li-Lee baseline at longer horizons. The parameter $\alpha$ is interpretable as the practitioner's confidence in the neural model's ability to extrapolate beyond the training period.

Together with the constraint intensity $\lambda$, the credibility parameter $\alpha$ defines a two-dimensional governance space. At one extreme ($\lambda \to 0$, $\alpha = 0$), the model is a pure unconstrained LSTM. At the other extreme ($\lambda \to \infty$, $\alpha \to \infty$), it collapses to the Li-Lee benchmark. Between these poles, the risk manager can select the operating point that best matches the regulatory context, the use case (best-estimate vs.\ prudent reserving), and the institutional risk appetite.


\section{Results}

\subsection{Ablation Studies: Hierarchy of Design Contributions}

To quantify the contribution of each design choice, we conducted systematic ablation experiments, progressively adding components to a minimal baseline and measuring the impact on out-of-sample RMSE. The results reveal a clear hierarchy of contributions (Table \ref{tab:ablation}).

\begin{table}[H]
\centering
\caption{Ablation Summary: Hierarchy of Design Contributions}
\label{tab:ablation}
\begin{tabular}{@{}lr@{}}
\toprule
Configuration & RMSE \\ \midrule
Separate M/F, fixed 32-16, lb=10 & 7.77 \\
+ Joint M/F ($+2.4\%$) & 7.59 \\
+ Architecture tuning via Keras Tuner ($+6.5\%$) & 7.07 \\
+ Lookback lb=15 via Optuna ($+12.7\%$) & 6.17 \\
+ Constraints $\lambda = 0.001$ ($-0.009\%$) & 6.17 \\
\bottomrule
\end{tabular}
\end{table}

The most important insight from the ablation is that temporal context dominates model capacity. The improvement from lookback optimization ($L = 10 \to 15$: $+12.7\%$) exceeds the improvement from architecture tuning ($+6.5\%$) by nearly a factor of two. This makes actuarial sense: mortality trends are driven by slow-moving structural factors (healthcare improvements, lifestyle changes, public health interventions) that unfold over decades. A model that sees 15 years of history captures the full arc of the post-2011 deceleration; a model with only 10 years of context sees only its tail. For practitioners, this suggests that investing in longer historical series (where data quality permits) may be more productive than investing in larger neural architectures.

The actuarial constraints are neutral on one-step RMSE ($-0.009\%$). This is a scientifically honest result that we do not attempt to overstate. The value of the constraints lies not in one-step accuracy but in multi-step forecast stability, as demonstrated in the constraint intensity experiment below.

\subsection{Constraint Intensity Trade-Off}\label{sec:constraint_tradeoff}

The central experiment of the paper. Five levels of $\lambda_{\text{monotonicity}}$ (0, 0.001, 0.01, 0.1, 1.0) were tested on the champion architecture with all other parameters fixed ($\lambda_{\text{coherence}} = 0.001$, residual-calibrated $\sigma$, 200 MC Dropout simulations). Each configuration was trained with seed 42 and its 30-year recursive forecast evaluated on four metrics: one-step RMSE, fraction of positive $\Delta K_t$ predictions, and 95\% confidence interval width for Male and Female life expectancy at the 2050 horizon.

\begin{table}[H]
\centering
\caption{Constraint Intensity Experiment (CHE, residual-calibrated $\sigma$)}
\label{tab:constraint}
\begin{tabular}{@{}rrrrrr@{}}
\toprule
$\lambda_{\text{mono}}$ & RMSE (cost) & frac($\Delta K_t > 0$) & CI$_M$ & CI$_F$ & CI Reduction \\ \midrule
0.0    & 6.173 (----)  & 55.6\% & 3.64 & 3.11 & ---    \\
0.001  & 6.173 ($-$0.01\%) & 55.6\% & 3.62 & 3.10 & $-$0.4\%  \\
0.01   & 6.171 ($-$0.03\%) & 55.6\% & 3.40 & 3.03 & $-$6.5\%  \\
0.1    & 6.225 ($+$0.84\%) & 44.4\% & 2.70 & 2.43 & $-$25.7\% \\
1.0    & 6.315 ($+$2.29\%) & 27.8\% & 1.93 & 1.71 & $-$46.8\% \\ \bottomrule
\end{tabular}
\end{table}

\begin{figure}[H]
    \centering
    \includegraphics[width=0.85\textwidth]{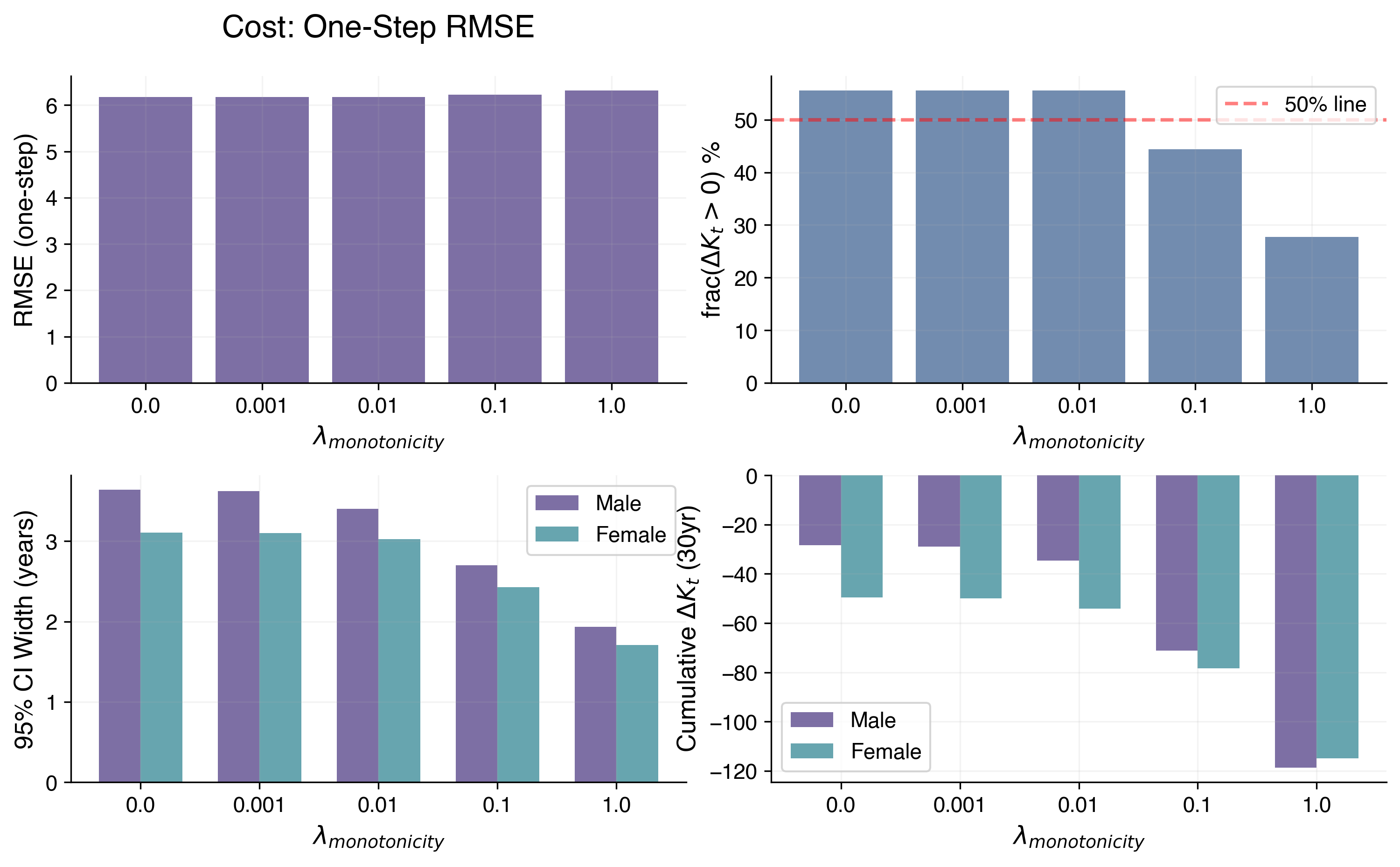}
    \caption{Constraint intensity trade-off across four metrics. The axes represent distinct concerns: one-step accuracy (top-left), biological compliance (top-right), and forecast stability for Male and Female (bottom). The ``sweet spot'' at $\lambda = 0.1$ reduces CI by 25\% at only $+0.84\%$ RMSE cost.}
    \label{fig:constraint}
\end{figure}

The results reveal a structured trade-off that admits three natural operating points, mapping directly to the actuarial model suite:

\textbf{$\lambda = 0.001$ (Optuna's champion): Best Estimate.} The constraint is too weak to affect either RMSE or CI. Optuna selects this value because it minimizes one-step validation error, and a $\lambda$ of 0.001 is indistinguishable from zero on a per-sample basis. This is the optimal choice for applications where point accuracy is paramount and governance is handled through other mechanisms (e.g., model passport, SHAP audit).

\textbf{$\lambda = 0.01$: Free Lunch.} Confidence intervals narrow by 6.5\% for males at \textit{zero RMSE cost} ($-0.03\%$). The constraint is strong enough to stabilize the recursive trajectory over 30 steps without being strong enough to degrade one-step predictions. This finding is invisible to standard hyperparameter optimizers, which evaluate $\lambda$ on one-step RMSE and would never discover the CI benefit. For practitioners, $\lambda = 0.01$ offers a strict improvement over the unconstrained model with no downside.

\textbf{$\lambda = 0.1$: Governance Sweet Spot.} CI narrows by 25\% at a cost of only $+0.84\%$ RMSE. The fraction of positive $\Delta K_t$ predictions drops from 55.6\% to 44.4\%, meaning the model is now biased toward mortality improvement (negative $\Delta K_t$) --- a bias that is biologically reasonable for developed countries over a 30-year horizon. This is the optimal choice for regulatory internal models where forecast stability and structural defensibility matter more than marginal one-step accuracy.

\textbf{$\lambda = 1.0$: Aggressive constraint.} CI is halved ($-47\%$) but the RMSE cost is substantial ($+2.29\%$) and the model becomes overly optimistic, projecting sustained mortality improvement that may be unrealistic. This level illustrates the upper bound of the trade-off but is not recommended for production use.

The disconnect between one-step RMSE (Optuna's target) and multi-step stability (the regulatory use case) is a general finding with implications beyond this specific model. It suggests that structural constraints in recursive neural forecasting should be evaluated on the target horizon, not on one-step validation --- a conclusion that applies to any neural time-series model used for long-term projection.

\subsection{Stochastic Life Expectancy Projections}

The ensemble with residual-calibrated $\sigma$ produces stochastic projections for 2020--2050. Table \ref{tab:e0} reports the full cluster results at the champion constraint level ($\lambda = 0.001$).

\begin{table}[H]
\centering
\caption{Ensemble Life Expectancy Projections 2020--2050 ($\lambda = 0.001$, residual-calibrated $\sigma$)}
\label{tab:e0}
\begin{tabular}{@{}llrrrr@{}}
\toprule
Country & Sex & $e_0$(2020) & $e_0$(2050) & CI Width & Gain \\ \midrule
Switzerland & M & 80.26 & 82.06 & 3.69 & +1.80 \\
Switzerland & F & 83.58 & 85.55 & 3.03 & +1.96 \\
Sweden & M & 79.94 & 81.74 & 3.73 & +1.80 \\
Sweden & F & 82.93 & 85.01 & 3.26 & +2.08 \\
Norway & M & 80.58 & 82.30 & 3.57 & +1.73 \\
Norway & F & 83.27 & 85.28 & 3.13 & +2.01 \\
W.\ Germany & M & 78.28 & 80.35 & 4.28 & +2.06 \\
W.\ Germany & F & 82.20 & 84.48 & 3.55 & +2.27 \\
Netherlands & M & 79.14 & 81.05 & 3.95 & +1.91 \\
Netherlands & F & 81.96 & 84.29 & 3.64 & +2.33 \\
Japan & M & 80.44 & 82.22 & 3.67 & +1.78 \\
Japan & F & 84.89 & 86.53 & 2.52 & +1.64 \\ \bottomrule
\end{tabular}
\end{table}

\begin{figure}[H]
    \centering
    \includegraphics[width=0.85\textwidth]{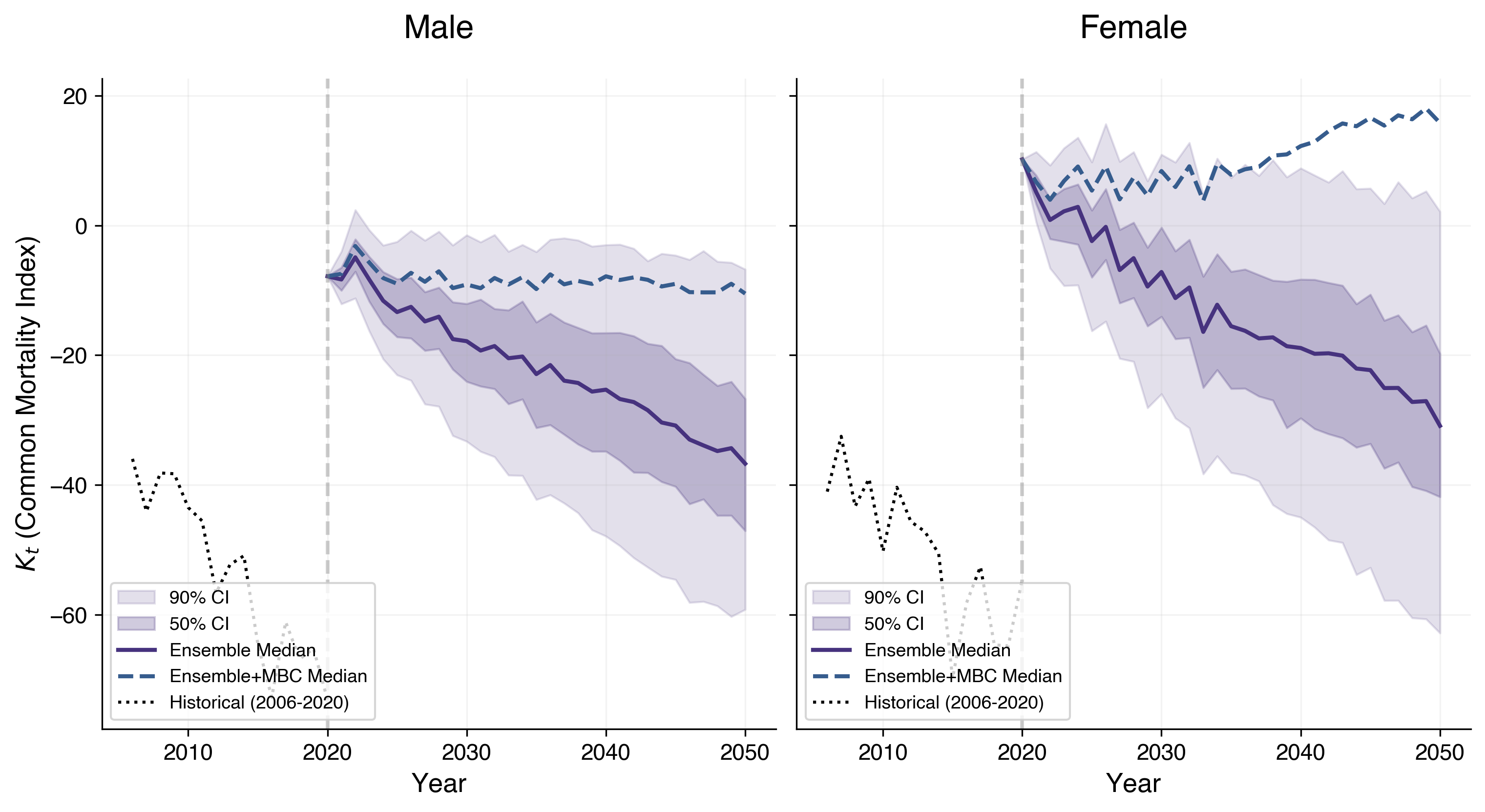}
    \caption{Stochastic $K_t$ fan charts for Male (left) and Female (right), 2020--2050. The 80\% and 95\% confidence intervals are derived from the 5-seed ensemble with residual-calibrated process noise. The Female fan chart is narrower despite a historically larger $\sigma$, reflecting the greater $\sigma$ reduction from residual calibration ($-60\%$ vs.\ $-53\%$).}
    \label{fig:fanchart}
\end{figure}

\begin{figure}[H]
    \centering
    \includegraphics[width=0.85\textwidth]{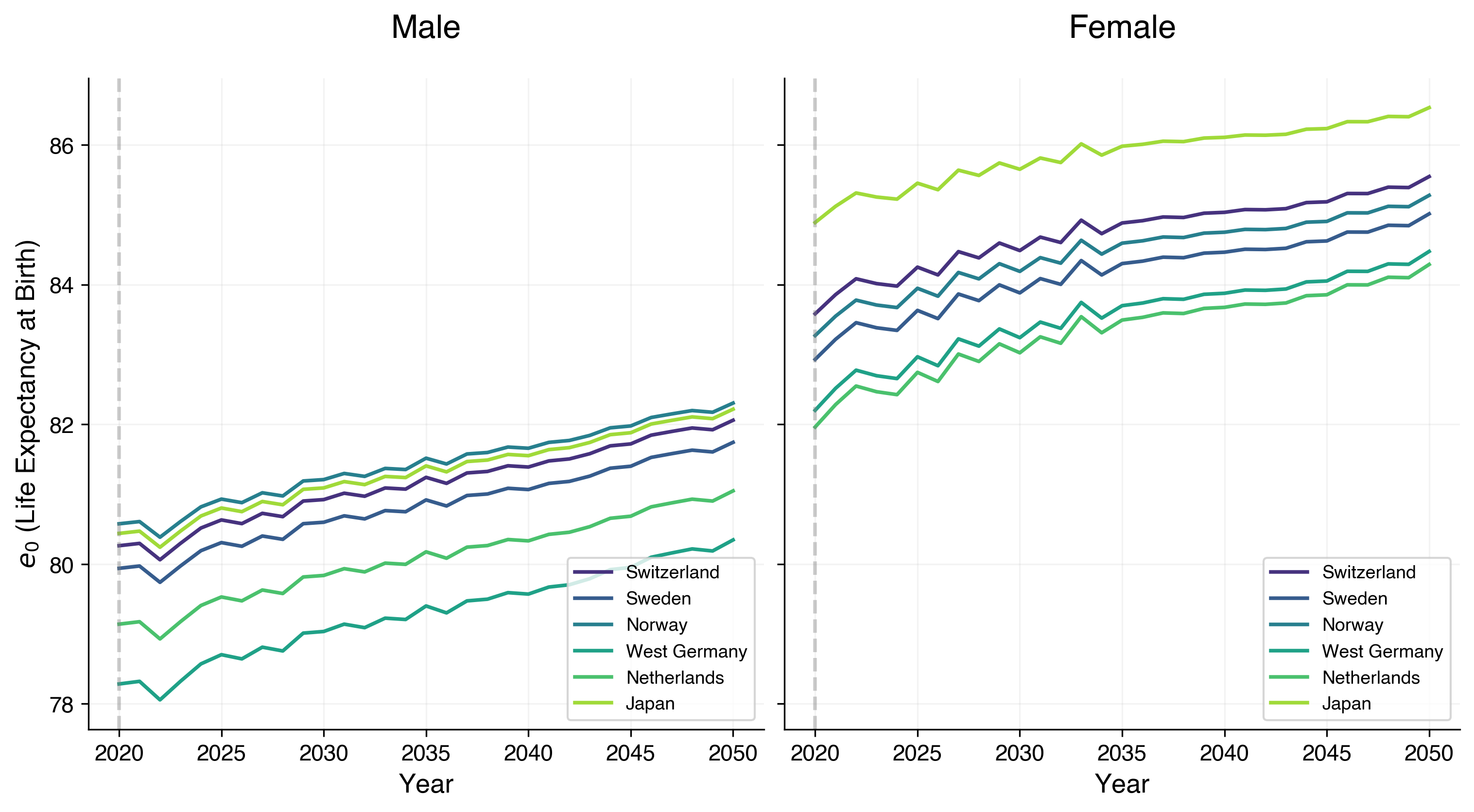}
    \caption{Life expectancy convergence across the cluster by sex. West Germany exhibits the steepest catch-up trajectory for both sexes, consistent with the persistent unit root in its country-specific factors. Japan Female has the smallest gain (+1.64 years) because it starts from the highest baseline (84.89), approaching a plausible biological ceiling.}
    \label{fig:convergence}
\end{figure}

Several patterns in the projections deserve discussion.

\textbf{Biological plausibility.} The projected gains ($+1.7$ to $+2.1$ years for males, $+1.6$ to $+2.3$ years for females over 30 years) are conservative compared to historical improvement rates but consistent with the post-2011 deceleration that the model has learned. The AINN with $L = 15$ has seen the full transition from fast improvement to deceleration and projects this slower regime forward --- a projection that is more prudent than the Li-Lee linear drift and more realistic than an unconstrained LSTM with shorter context.

\textbf{Sex differential.} Female gains are consistently larger than male gains ($+0.2$ to $+0.5$ years more) for most countries. This reflects the model's learning that female mortality still has more room for improvement in the current regime, consistent with the steeper female $K_t$ decline. The exception is Japan, where the female gain ($+1.64$) is the smallest in the cluster because Japan Female starts from the highest baseline ($e_0 = 84.89$) --- approaching a plausible biological ceiling beyond which further gains become increasingly difficult.

\textbf{Catch-up dynamics.} West Germany shows the largest male gain ($+2.06$) and the Netherlands the largest female gain ($+2.33$), consistent with the catch-up dynamics identified in \cite{rindori2026p04}. These are the countries with the strongest unit roots in their specific factors --- persistent structural drifts that the Li-Lee framework suppresses but the LSTM can internalize through the joint multi-population training.

\textbf{CI asymmetry.} Female confidence intervals are narrower than male for all countries except the Netherlands. This is a direct consequence of the residual calibration: the 60\% reduction in female $\sigma$ (vs.\ 53\% for males) more than compensates for the historically higher female variability, resulting in tighter female projections.

\subsection{Benchmark: AINN vs.\ Li-Lee RWD}

To contextualize the AINN's projections against the actuarial benchmark, we compare them with sex-specific Li-Lee Random Walk with Drift (RWD) projections. The Li-Lee RWD is the standard actuarial production model: $K_t$ follows a linear drift estimated from the full historical series, and specific factors are frozen at their last observed values.

\begin{table}[H]
\centering
\caption{AINN Ensemble vs.\ Li-Lee RWD --- Switzerland at 2050}
\label{tab:benchmark}
\begin{tabular}{@{}llrrrr@{}}
\toprule
Sex & Model & $e_0$(2050) & Gain & CI Width & SCR (ES) \\ \midrule
M & Li-Lee & 83.54 & +3.28 & 3.73 & +2.091 \\
M & AINN   & 82.06 & +1.80 & 3.69 & +1.907 \\
F & Li-Lee & 86.05 & +2.47 & 4.71 & +2.178 \\
F & AINN   & 85.55 & +1.96 & 3.03 & +1.516 \\ \bottomrule
\end{tabular}
\end{table}

\begin{figure}[H]
    \centering
    \includegraphics[width=0.85\textwidth]{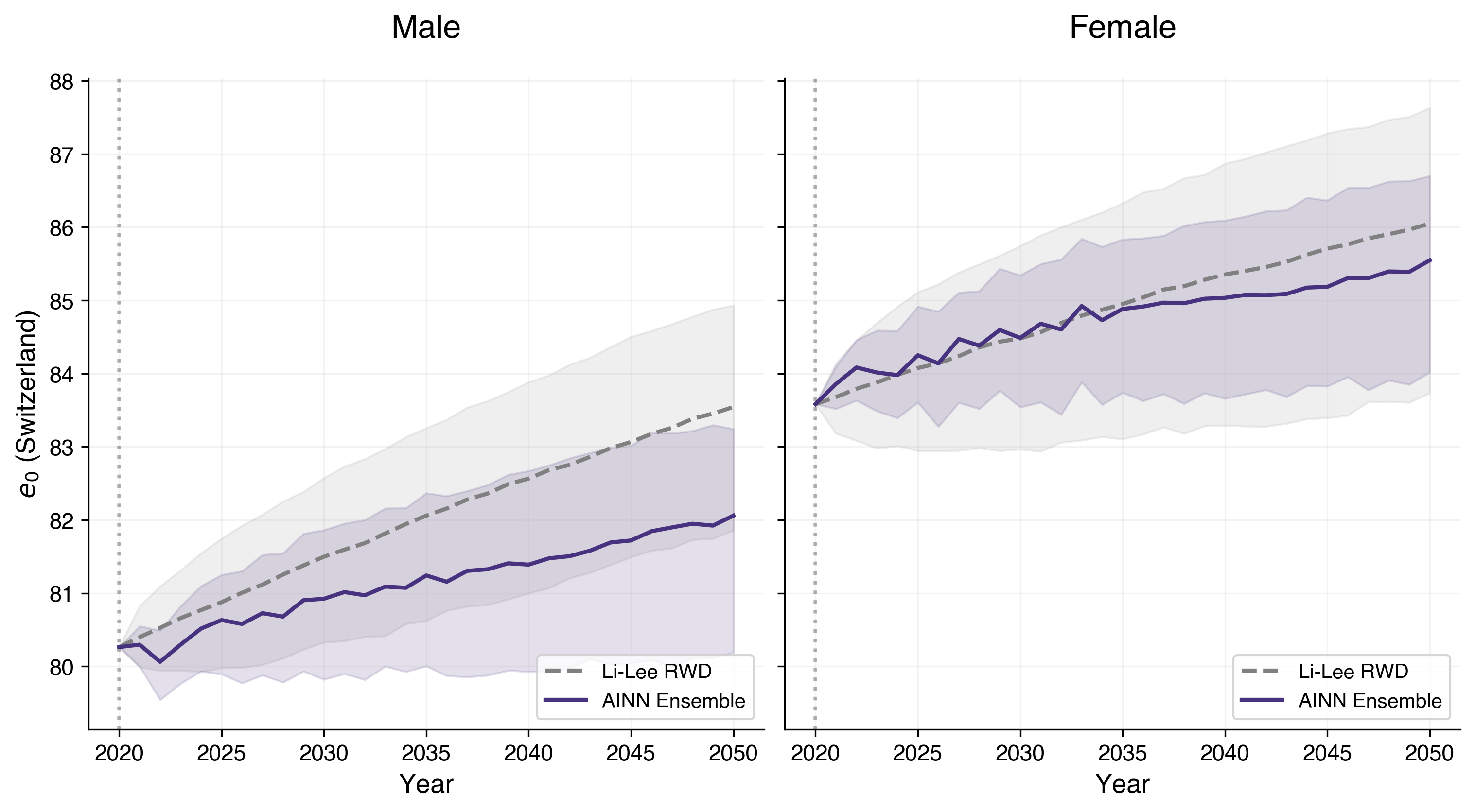}
    \caption{Switzerland $e_0$: AINN ensemble (colored, with CI) vs.\ Li-Lee RWD (grey). The AINN captures the post-2011 deceleration, producing more conservative median projections. The advantage is most visible for females, where the AINN's residual-calibrated CI is 36\% narrower than Li-Lee's.}
    \label{fig:benchmark}
\end{figure}

The comparison reveals two distinct advantages. First, the AINN's median projections are more conservative ($+1.80$ vs.\ $+3.28$ for males, $+1.96$ vs.\ $+2.47$ for females). This conservatism is not a limitation but a feature: the model has learned the post-2011 deceleration and projects it forward, whereas the Li-Lee drift is estimated over the full 1956--2020 period and is dominated by the faster improvement of the pre-2011 era. For reserving purposes, the AINN's projection is more prudent.

Second, the AINN's confidence intervals are narrower --- particularly for females, where the CI width is 3.03 vs.\ 4.71 years ($-36\%$). This is the combined effect of residual-calibrated $\sigma$ (which eliminates double-counting) and the 5-seed ensemble (which reduces seed-related variance). Narrower CI translates directly to lower regulatory capital requirements: the AINN's SCR for Switzerland Female is $+1.516$ years vs.\ $+2.178$ for Li-Lee ($-30\%$). For a pension fund or reinsurer, this difference in capital requirement is economically significant.

\subsection{10-Year Out-of-Sample Backtest}

To validate the framework on truly unseen data, we retrained the AINN on the 1956--2010 period only and forecasted 2011--2020, a decade that includes both the post-2011 mortality improvement deceleration and the 2020 COVID shock. This is the most demanding test: the model must extrapolate beyond the regime it was trained on.

\begin{table}[H]
\centering
\caption{10-Year Backtest: Predicted vs.\ Observed $e_0$ at 2020 (Selected Countries)}
\label{tab:backtest}
\begin{tabular}{@{}llrrrll@{}}
\toprule
Country & Sex & Observed & AINN & Li-Lee & AINN Error & In 90\% CI \\ \midrule
CHE & M & 80.26 & 80.52 & 80.51 & $+$0.26 & YES \\
CHE & F & 83.58 & 83.84 & 84.04 & $+$0.26 & YES \\
NOR & M & 80.58 & 79.56 & 79.54 & $-$1.02 & NO \\
JPN & F & 84.89 & 84.78 & 84.95 & $-$0.11 & YES \\ \bottomrule
\end{tabular}

\vspace{0.2cm}
\footnotesize{Full cluster: AINN MAE = 0.379 years, Li-Lee MAE = 0.427 years. AINN wins 6/12 country$\times$sex comparisons. 11/12 within the 90\% CI.}
\end{table}

\begin{figure}[H]
    \centering
    \includegraphics[width=0.85\textwidth]{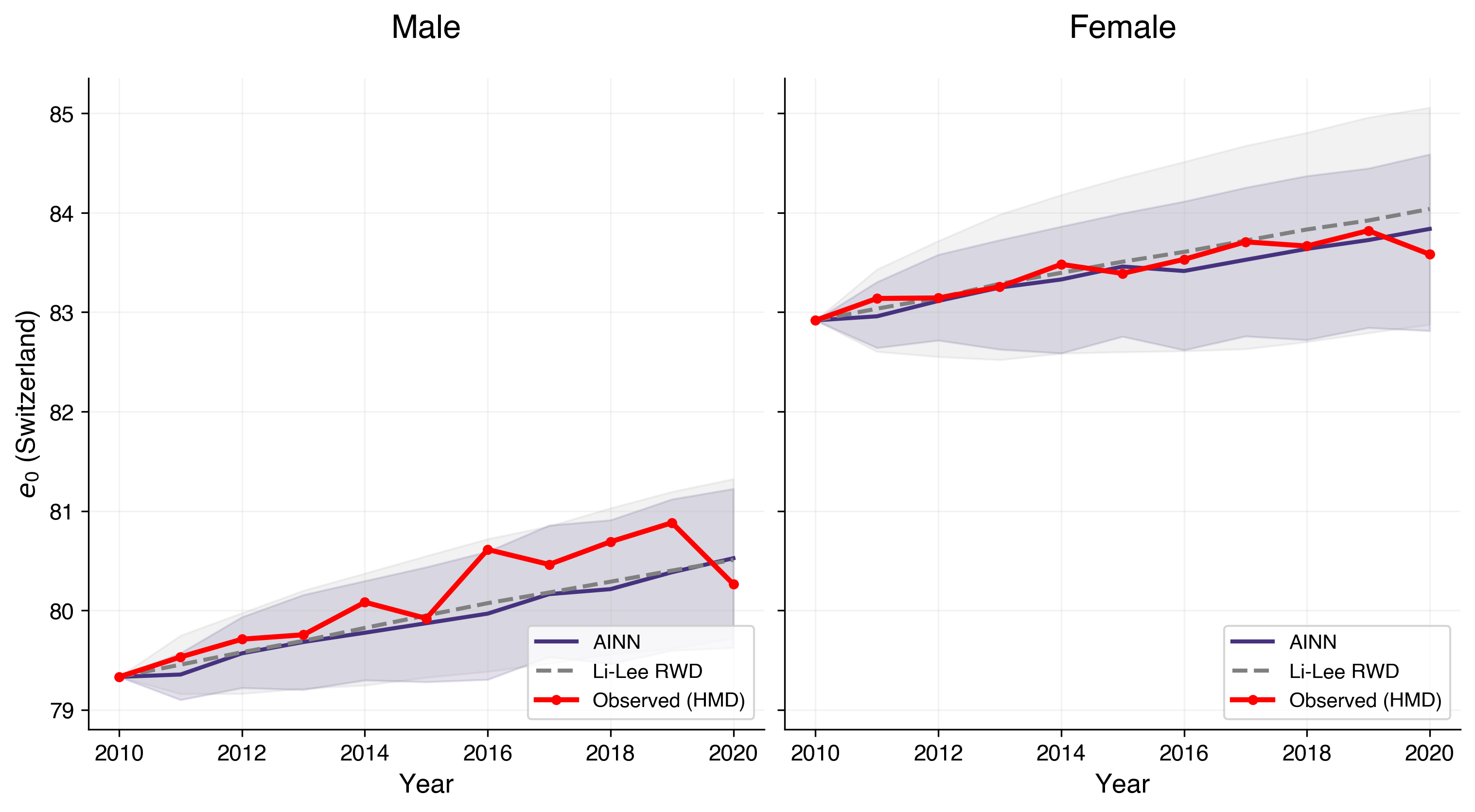}
    \caption{10-year backtest for Switzerland: predicted vs.\ observed $e_0$ (red circles). Both models track the trajectory reasonably well; the AINN's advantage is in uncertainty calibration (narrower CI for females) rather than in point accuracy.}
    \label{fig:backtest}
\end{figure}

The backtest shows that the AINN does not dramatically outperform Li-Lee on point accuracy (MAE 0.379 vs.\ 0.427 years). This is an honest result: on a 10-year horizon, where the underlying trend is approximately linear, the Li-Lee's linear drift is a reasonable approximation. The AINN's advantage lies in uncertainty calibration, not in point prediction. For regulatory applications where CI width directly affects capital requirements, the AINN's tighter intervals (particularly for females) represent a material improvement. For point-estimate-only applications, the AINN's additional complexity may not be justified --- a conclusion we state transparently.

The single failure (Norway Male, outside the 90\% CI) is attributable to Norway's small-population volatility, which produces higher variance in all models. Norway has consistently been the most challenging population in the cluster for both \cite{rindori2026p04} and this work.

\subsection{Explainability and Governance}

\subsubsection{Temporal Saliency}

Gradient-based sensitivity analysis quantifies which years in the 15-year input window drive the forecast. The results show a distributed importance pattern: Male peaks at $t-3$ (9.9\%), Female at $t-4$ (10.2\%), with no single lag dominating. All lags between $t-3$ and $t-10$ contribute between 6\% and 10\%.

\begin{figure}[H]
    \centering
    \includegraphics[width=0.85\textwidth]{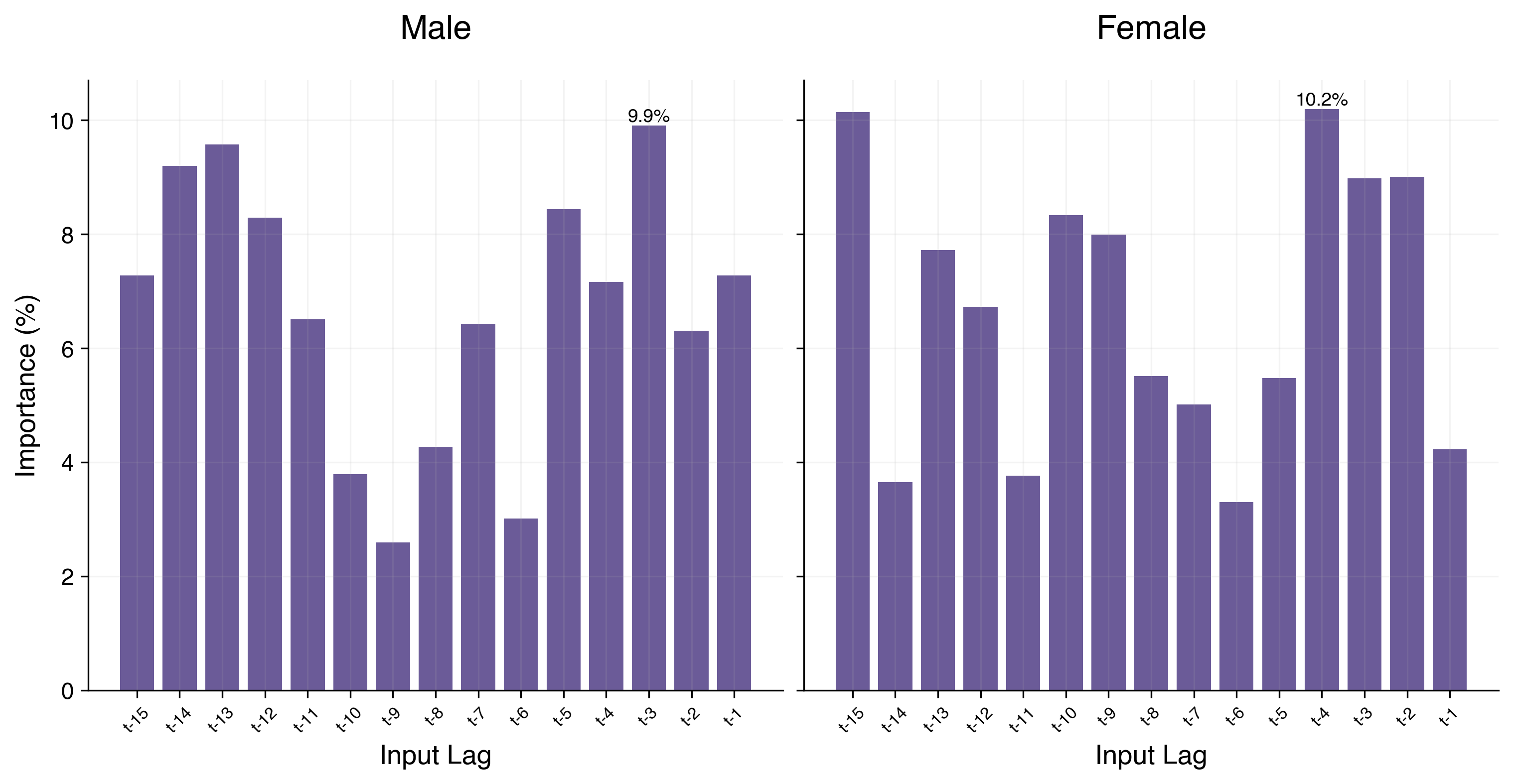}
    \caption{Temporal saliency profiles for Male and Female. The distributed importance across the full 15-year window contrasts with the concentrated $t-4$ peak (21.7\%) observed in \cite{rindori2026p04} with $L = 10$, suggesting that the longer window allows the model to integrate evidence more broadly.}
    \label{fig:saliency}
\end{figure}

This distributed pattern contrasts markedly with the concentrated $t-4$ peak (21.7\%) found in \cite{rindori2026p04} with $L = 10$. The difference is a direct consequence of the longer lookback window: with 15 years of context, the model can afford to distribute attention broadly rather than concentrating on a single ``pivotal'' lag. From a regulatory perspective, this is a desirable property --- a model that relies on the full history is more robust than one that depends critically on a single observation.

\subsubsection{SHAP Influence Mapping}

SHAP (KernelExplainer) decomposes the Swiss Male mortality forecast into contributions from each input factor. The ranking reveals Netherlands (0.063) as the top external influence, followed by Japan (0.040), Sweden (0.038), and Switzerland's own autoregressive signal (0.037).

\begin{figure}[H]
    \centering
    \includegraphics[width=0.85\textwidth]{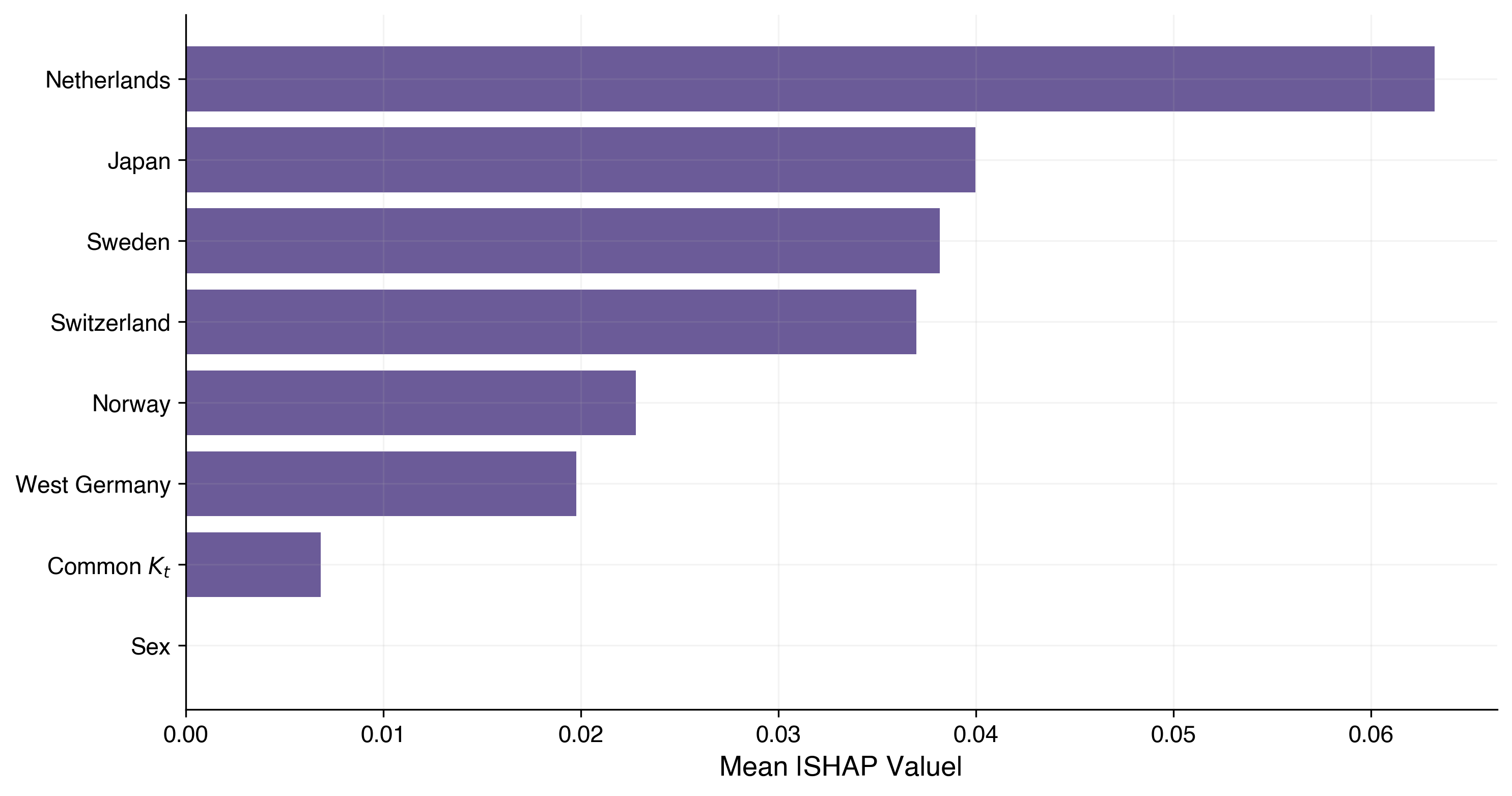}
    \caption{SHAP influence ranking for Swiss Male mortality. The distributed pattern across all cluster members indicates that the LSTM integrates the full multi-population structure. The common factor $K_t$ ranks lowest because its marginal contribution over the background is minimal (all samples share a similar $K_t$ pattern).}
    \label{fig:shap}
\end{figure}

The SHAP hierarchy differs from \cite{rindori2026p04}, where Norway was the top predictor. This difference is attributable to the different architecture ($48 \to 32$ vs.\ $32 \to 16$), lookback ($L = 15$ vs.\ $L = 10$), and the sex-specific decomposition, which changes the information content of each input feature. The common factor $K_t$ ranks lowest among mortality factors --- counterintuitive at first glance, but expected: SHAP measures \textit{marginal} contribution relative to the background distribution. Since all samples share a similar $K_t$ pattern, its marginal variation is low; the country-specific factors vary more across samples and therefore receive higher SHAP attribution.

\subsubsection{Biological Consistency: Gompertz Audit}\label{sec:gompertz}

The projected mortality curves (ages 40--90, 2050) pass the Gompertz monotonicity test: mortality increases strictly with age for all country$\times$sex combinations after isotonic correction. Small violations ($< 0.001$) exist in the raw projections but are inherited from HMD data granularity artefacts, not generated by the model. The observation-anchored approach (Equation \ref{eq:anchored}) applies a uniform shift ($B_x \cdot \Delta K_t$) that mathematically cannot introduce new monotonicity violations.

\subsubsection{Lexis Residual Maps}

\begin{figure}[H]
    \centering
    \includegraphics[width=0.85\textwidth]{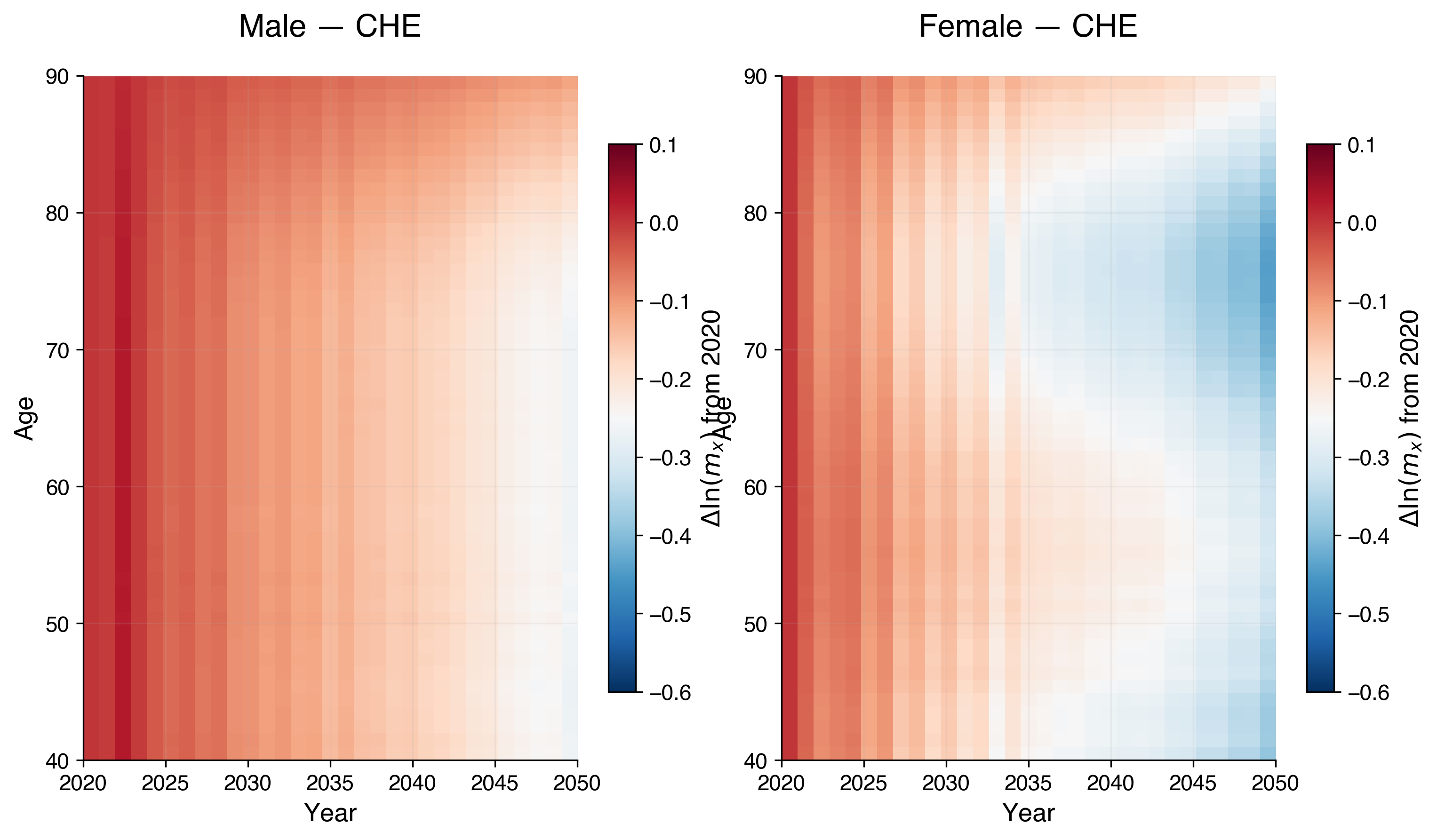}
    \caption{Lexis residual maps for Switzerland (ages 40--90). The smooth vertical gradient with no diagonal cohort artefacts confirms that the rank-1 reconstruction captures the dominant age-period interaction without introducing spurious cohort effects.}
    \label{fig:lexis}
\end{figure}

The Lexis residual maps show smooth vertical gradients with no diagonal cohort artefacts, confirming that the rank-1 observation-anchored reconstruction captures the dominant age-period interaction cleanly.

\subsection{Regulatory Capital and Stress Testing}

\subsubsection{Solvency Capital Requirement}

\begin{table}[H]
\centering
\caption{SCR (ES 99.0\%, SST) --- Full Cluster}
\label{tab:scr}
\begin{tabular}{@{}lrr@{}}
\toprule
Country & Male SCR & Female SCR \\ \midrule
Switzerland  & +1.907 & +1.516 \\
Sweden       & +1.942 & +1.649 \\
Norway       & +1.852 & +1.580 \\
W.\ Germany  & +2.245 & +1.800 \\
Netherlands  & +2.063 & +1.849 \\
Japan        & +1.902 & +1.243 \\ \bottomrule
\end{tabular}
\end{table}

\begin{figure}[H]
    \centering
    \includegraphics[width=0.85\textwidth]{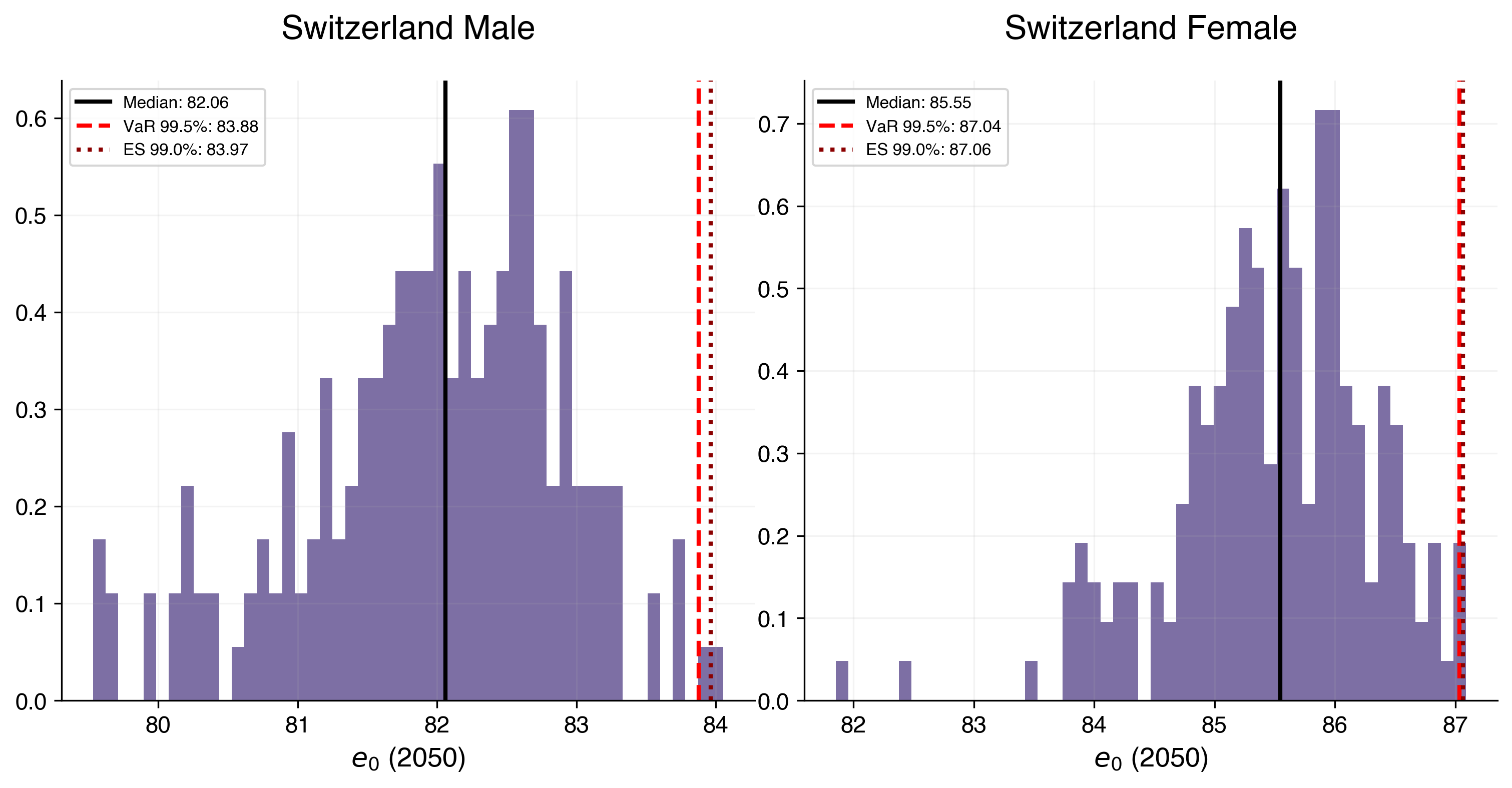}
    \caption{Tail risk distribution for Swiss $e_0$ at 2050 (Male and Female). The ES 99.0\% thresholds are indicated, defining the SCR under the SST framework.}
    \label{fig:scr}
\end{figure}

Male SCR exceeds Female SCR for all countries, reflecting the higher residual uncertainty in male mortality dynamics. West Germany carries the highest tail risk for both sexes ($+2.245$ M, $+1.800$ F), consistent with its catch-up volatility. Japan Female has the lowest SCR ($+1.243$), as it starts from the highest baseline and has the least room for surprise longevity gains.

\subsubsection{Reverse Stress Test}

Following the FINMA methodology, we compute the critical mortality reduction $\delta^*$ that would exhaust the ES 99.0\% solvency buffer. For Switzerland: $\delta^* = 23.0\%$ (Male), $25.2\%$ (Female). The linearity of the $\delta \to \Delta e_0$ mapping is verified ($R^2 > 0.999$). A standard 10\% shock consumes 39--44\% of the SCR buffer.

\subsubsection{Model-Based Stress Test}

Unlike the level-shift reverse stress test, which applies a post-hoc shock to the mortality surface, the model-based stress test translates a mortality shock into the $\Delta K_t$ domain and injects it into the LSTM's sliding window at the shock year (2026). The recursive forecast is then re-executed with MC Dropout, allowing the model's dynamic response --- including potential non-linear amplification or dampening --- to be observed directly.

\begin{figure}[H]
    \centering
    \includegraphics[width=0.85\textwidth]{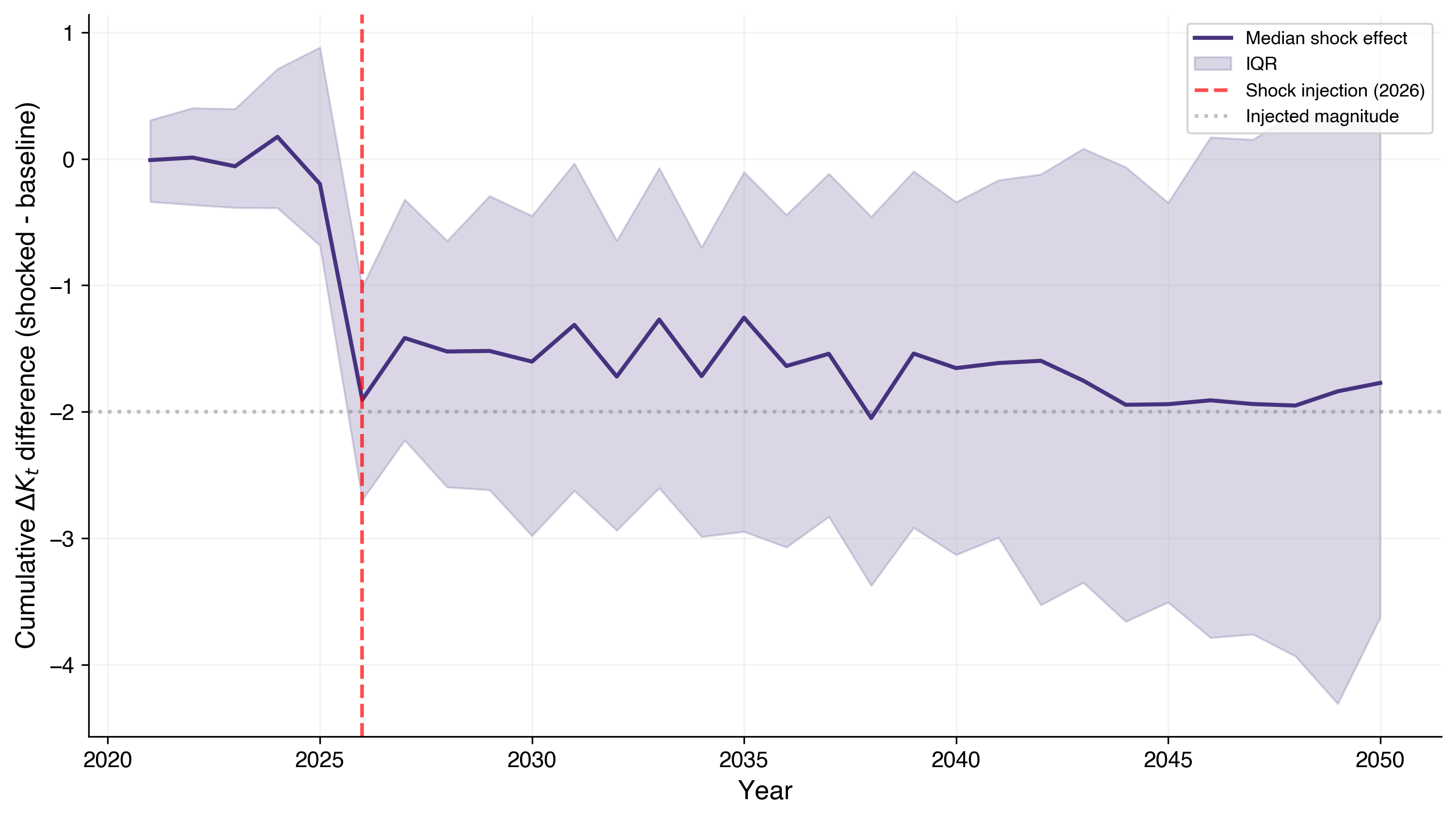}
    \caption{Model-based stress test response. A negative shock of $-2\sigma$ in $\Delta K_t$ space is injected at 2026. The model absorbs the shock without amplification (maximum ratio 1.02$\times$), confirming numerical stability under perturbation.}
    \label{fig:stress}
\end{figure}

The result is unambiguous: the maximum amplification ratio is $1.02\times$ --- the model is \textbf{STABLE}. The shock is absorbed smoothly; the effect stabilizes within the lookback window without explosive feedback. This confirms that the AINN can be safely used for scenario analysis without producing unrealistic cascading effects --- a key requirement for FINMA internal model validation.

\subsection{Credibility-Weighted MBC: The 2D Governance Space}

\begin{table}[H]
\centering
\caption{Credibility $Z(t) = e^{-\alpha t}$: Effect on CHE Projections}
\label{tab:credibility}
\begin{tabular}{@{}rrrrrrr@{}}
\toprule
$\alpha$ & $Z$(2025) & $Z$(2035) & $Z$(2050) & M Med & F Med & M CI \\ \midrule
0.0  & 1.000 & 1.000 & 1.000 & 82.06 & 85.55 & 3.69 \\
0.02 & 0.905 & 0.741 & 0.549 & 81.69 & 85.14 & 2.89 \\
0.05 & 0.779 & 0.472 & 0.223 & 81.26 & 84.73 & 2.16 \\
0.10 & 0.607 & 0.223 & 0.050 & 80.90 & 84.33 & 1.81 \\
0.20 & 0.368 & 0.050 & 0.002 & 80.63 & 83.98 & 1.35 \\ \bottomrule
\end{tabular}
\end{table}

The table illustrates the continuous nature of the blending. At $\alpha = 0.02$, the model retains 90\% credibility at 5 years but only 55\% at 30 years, producing projections that track the LSTM closely in the near term but gradually revert to the Li-Lee baseline. At $\alpha = 0.20$, the credibility collapses to near-zero beyond 15 years, and the projection is essentially the Li-Lee drift.

Together, the constraint intensity $\lambda$ and the credibility parameter $\alpha$ define a two-dimensional governance space. The $\lambda$-axis controls structural compliance (how much the training process respects actuarial priors). The $\alpha$-axis controls temporal trust (how far into the future the neural signal is credible). A risk manager can select the point $(\lambda, \alpha)$ that matches the regulatory context: a best-estimate for pricing ($\lambda = 0.001$, $\alpha = 0$), a prudent reserve ($\lambda = 0.1$, $\alpha = 0.05$), or a stress scenario ($\lambda = 1.0$, $\alpha = 0.10$).


\section{Discussion}

\subsection{The Constraint as a Governance Dial}

The AINN's primary contribution is making explicit a trade-off that is hidden in both classical and neural models. The Li-Lee framework hardcodes its structural assumptions: coherence is enforced by construction, and there is no knob to turn if the practitioner wants more or less flexibility. An unconstrained LSTM has no structural bias whatsoever --- it learns whatever the data contain, including noise and artefacts. Neither extreme offers the risk manager a choice.

The AINN provides this choice through the $\lambda$ parameter. The constraint intensity experiment (Section \ref{sec:constraint_tradeoff}) demonstrates that this is not a theoretical abstraction but a practically useful instrument: at $\lambda = 0.01$, the model gains 6.5\% tighter confidence intervals at zero accuracy cost, while at $\lambda = 0.1$, confidence intervals tighten by 25\% at a cost that most practitioners would consider negligible ($+0.84\%$ RMSE). The ability to dial this trade-off continuously --- rather than making a binary ``constrained or unconstrained'' choice --- is the operational innovation.

A key implication is that the \textit{right} value of $\lambda$ depends on the use case, not on the data. For a best-estimate projection used in pricing, $\lambda = 0.001$ (maximum accuracy) is appropriate. For a prudent reserve used in regulatory reporting, $\lambda = 0.1$ (stability over accuracy) is preferable. For a stress scenario, $\lambda = 1.0$ (maximum constraint) may be warranted. The AINN framework does not prescribe a single answer; it provides the mechanism for the risk manager to make an informed choice.

\subsection{Residual-Calibrated Uncertainty: The Largest Single Improvement}

The 53--60\% reduction in process noise $\sigma$ is the largest single improvement in the pipeline, exceeding the impact of the ensemble ($\approx$15\% CI reduction from averaging), the constraint ($\leq$25\% at $\lambda = 0.1$), and the lookback extension ($\approx$13\% RMSE). This finding has implications beyond the specific AINN model: it suggests that in any dual uncertainty framework that combines model predictions with additive process noise, the calibration of the aleatoric component deserves at least as much attention as the design of the model itself.

The principle is straightforward: if the model has learned to predict part of the historical variability, then using the full historical variability as process noise counts that part twice. The residual-based calibration is the natural correction. We note that the original pipeline in \cite{rindori2026p04} also contained this double-counting, but it was less visible because the combined-sex $\sigma$ was smaller and the resulting CI ($\pm 0.9$ years) fell within a plausible range. The sex-specific decomposition amplified the problem to the point where it became obvious ($\pm 6.4$ years), prompting the correction. This is an instructive example of how model extensions can expose latent issues in the original framework.

\subsection{Honest Assessment: When Is the AINN Justified?}

The 10-year backtest provides an honest basis for answering this question. On point accuracy alone, the AINN does not dramatically outperform Li-Lee (MAE 0.379 vs.\ 0.427 years). The advantage is in uncertainty calibration: tighter confidence intervals, particularly for females ($-36\%$ CI width), which translate to lower regulatory capital requirements ($-30\%$ SCR for Swiss females). For applications where only point estimates matter, the AINN's additional complexity is difficult to justify. For regulatory applications where CI width directly affects capital, the improvement is economically significant.

This assessment should inform adoption decisions. The AINN is most valuable as a challenger model in the internal model validation suite --- an independent, methodologically distinct projection that can be compared against the Li-Lee production model to identify blind spots, regime changes, and calibration issues. It is not a replacement for Li-Lee: the classical model's simplicity, transparency, and long track record remain compelling advantages for primary production use.

\subsection{Sex-Dependent Stationarity: Implications for the Literature}

The finding that sex-specific decomposition changes the stationarity classifications of country-specific factors (Section \ref{sec:stationarity}) has implications beyond this specific model. Sweden's reversal from ``Unit Root'' to ``Stationary'' upon sex disaggregation demonstrates that combined-sex analysis can introduce spurious non-stationarity through an aggregation artefact. This suggests that multi-population mortality studies should routinely report sex-specific stationarity results alongside combined results, as the choice of aggregation level can qualitatively change the assessment of a model's foundational assumptions.

\subsection{Limitations}

Several limitations should be acknowledged transparently.

\textbf{Constraint effect at champion $\lambda$.} At $\lambda = 0.001$ (the value selected by Optuna), the constraint has near-zero effect on both RMSE and confidence intervals. The governance-accuracy trade-off emerges only through post-hoc analysis at higher $\lambda$ values that the optimizer would not choose. A multi-step training objective that directly penalizes 30-year forecast variance would provide a principled basis for $\lambda$ selection, but would require fundamental changes to the training loop.

\textbf{Temporal monotonicity surrogate.} The monotonicity penalty $\text{ReLU}(\Delta K_t)^2$ penalizes mortality worsening over time, but does not enforce the true Gompertz constraint ($m_{x+1} \geq m_x$ for all ages). Implementing the true age-monotonicity constraint in the loss requires differentiable back-transformation from $\Delta K_t$ to age-specific death rates --- computationally expensive and numerically fragile for a rank-1 reconstruction. The current approach (temporal surrogate + post-hoc isotonic correction) is pragmatic and defensible, but the theoretically correct alternative remains future work.

\textbf{Credibility parameter $\alpha$.} The decay rate $\alpha$ in the time-varying MBC has no data-driven calibration in this work. It is presented as an interpretable governance parameter, but a future extension could derive $\alpha$ from the model's rolling forecast variance, providing a self-calibrating credibility weight.

\textbf{Small dataset.} The training set contains 80 sequences (40 per sex with $L = 15$). This is fundamental to mortality modeling --- annual data from 65 years of observation cannot be augmented without synthetic generation. The 5-seed ensemble and the rolling-window validation partially mitigate this limitation, but it remains the binding constraint on model complexity.

\textbf{Aggregation artefact.} The finding that sex-specific decomposition changes stationarity classifications (Section \ref{sec:stationarity}) nuances the motivating narrative from \cite{rindori2026p04}. The ``stationarity paradox'' identified there was partly an artefact of combined-sex analysis. The sex-specific analysis reveals a less dramatic but more nuanced picture where the stationarity landscape depends on the aggregation level.

\textbf{Female RMSE exceeds Male RMSE} (6.59 vs.\ 5.72), a gap that persists across all configurations. Female mortality is intrinsically harder to predict in this cluster, consistent with the higher volatility of female-specific factors and the more complex interaction between biological and behavioral drivers of female longevity.


\section{Conclusion}

This paper introduced Actuarial-Informed Neural Networks (AINN), a constrained deep learning framework for sex-specific multi-population longevity forecasting. The framework adapts the Physics-Informed Neural Networks paradigm to actuarial science, embedding Li-Lee coherence and temporal monotonicity as differentiable penalty terms in the LSTM training loss. We applied it to a frontier cluster of six high-longevity nations (Switzerland, Sweden, Norway, West Germany, Netherlands, Japan; 1956--2020), positioning the AINN as a challenger model to both the Li-Lee benchmark and unconstrained neural approaches.

The central finding is that the constraint intensity $\lambda$ functions as a continuous governance dial with a quantifiable accuracy-stability trade-off. At $\lambda = 0.01$, the constraint is effectively free: confidence intervals narrow by 6.5\% at zero RMSE cost. At $\lambda = 0.1$, they narrow by 25\% at a cost of only $+0.84\%$ RMSE. This trade-off is invisible to standard hyperparameter optimizers, which evaluate $\lambda$ on one-step accuracy rather than multi-step stability --- a general finding that applies to any structural constraint in recursive neural forecasting.

Three additional contributions complete the framework. First, residual-calibrated process noise eliminates the double-counting problem in dual uncertainty frameworks, reducing 95\% confidence intervals by 53--60\% --- the single largest improvement in the pipeline. Second, joint Male/Female training with a 5-seed ensemble produces coherent sex-specific projections from a single model (CV = 1.06\%). Third, a time-varying credibility weight $Z(t) = e^{-\alpha t}$ connects the Mean-Bias Correction to B\"uhlmann credibility theory, providing a principled framework for blending neural and actuarial predictions across forecast horizons.

A 10-year out-of-sample backtest on truly unseen data (train 1956--2010, validate 2011--2020) yields a mean absolute error of 0.38 years, with 11 of 12 country$\times$sex combinations within the 90\% confidence interval. The framework produces regulatory-grade outputs: SCR under both SST (ES 99.0\%: $+1.91$ years, CHE Male) and Solvency II standards, reverse stress testing ($\delta^* = 23.0\%$, CHE Male), and model-based stability verification (amplification ratio $1.02\times$).

The AINN does not replace classical models. It extends them with tunable constraints and calibrated uncertainty, filling the gap between Li-Lee and standard neural networks in a regulatory environment where governance matters as much as accuracy. The $(\lambda, \alpha)$ governance space provides a single, coherent framework from which a risk manager can derive the full actuarial model suite --- best estimate, prudent reserve, and stress scenario --- by adjusting two interpretable parameters.

\subsection{Future Work}

Several extensions follow naturally from this work. A multi-step training objective that directly optimizes for 30-year forecast stability would provide a principled basis for $\lambda$ selection, replacing the current post-hoc analysis. Fat-tailed process noise (Student-$t$ or bootstrap residuals) would better capture extreme demographic events. The true Gompertzian constraint ($m_{x+1} \geq m_x$ for all ages) in the loss function, requiring differentiable back-transformation, would replace the temporal surrogate with a structurally exact alternative. A self-calibrating credibility parameter $\alpha(t)$ derived from rolling forecast variance would eliminate the need for practitioner specification. Finally, diagonal cohort features in the input tensor could capture cohort effects that the current age-period decomposition misses.

\section*{Acknowledgements}
The author thanks Riccardo Turin (Actuary for Pricing Tools, Swiss Re) for feedback on \cite{rindori2026p04} that informed this work.

\newpage
\bibliographystyle{unsrt}
\bibliography{references}

\end{document}